\documentclass[lettersize,journal]{IEEEtran}
\usepackage{amsmath,amsfonts}
\usepackage{stfloats}
\usepackage{url}
\usepackage{verbatim}
\usepackage{graphicx}
\usepackage{cite}
\usepackage{array}
\usepackage[lined,boxed,commentsnumbered, ruled]{algorithm2e}
\usepackage{textcomp}
\usepackage{bbm}
\usepackage{mathrsfs}
\usepackage{bm}

\usepackage{booktabs,multirow}
\usepackage{enumerate} 
\usepackage{epstopdf}
\usepackage{tikz}
\usetikzlibrary{arrows,decorations.markings}
\usepackage{setspace}
\usepackage{graphics}
\usepackage[labelformat=simple]{subcaption}
\usetikzlibrary{shapes}
\usepackage{amsthm}
\definecolor{colorhkust}{HTML}{142B8C}
\definecolor{colorshanghaitech}{HTML}{A20005}
\definecolor{colortsinghua}{HTML}{743481}
\definecolor{colordark}{RGB}{184,134,11}
\definecolor{colorRed}{RGB}{128, 0, 0}
\definecolor{colorGreen}{RGB}{0, 64, 0}
\definecolor{colorBlue}{RGB}{0, 0, 128}
\usepackage[
colorlinks, 
linkcolor=colorRed, 
anchorcolor=blue, 
citecolor=colorBlue, 
urlcolor = colorGreen
]{hyperref}
\usepackage{etoolbox}
\usepackage{letltxmacro}
\usepackage[compatibility=false]{caption}

\begin{document}

\title{Satellite Federated Fine-Tuning for Foundation Models in Space Computing Power Networks}
\author{Yan~Zhu,~\IEEEmembership{Graduate Student Member, IEEE,}
        Jingyang~Zhu,~\IEEEmembership{Graduate Student Member, IEEE,}\\
        Ting~Wang,~\IEEEmembership{Senior Member, IEEE,}
        Yuanming~Shi,~\IEEEmembership{Senior Member, IEEE,}
        Chunxiao~Jiang,~\IEEEmembership{Fellow, IEEE,}
        and~Khaled B.~Letaief,~\IEEEmembership{Fellow, IEEE}
\thanks{Yan Zhu, and Ting Wang are with the MoE Engineering ResearchCenter of Software/Hardware Co-design Technology and Application, the Shanghai Key Lab. of Trustworthy Computing, East China Normal University, Shanghai 200062, China (e-mail: 51275902041@stu.ecnu.edu.cn; twang@sei.ecnu.edu.cn).

Yuanming Shi and Jingyang Zhu are with the School of Information Science and Technology, ShanghaiTech University, Shanghai 201210, China (e-mail: shiym@shanghaitech.edu.cn; zhujy2@shanghaitech.edu.cn). 

Chunxiao Jiang is with Beijing National Research Center for Information Science and Technology, Tsinghua University, Beijing 100084, China (e-mail: jchx@tsinghua.edu.cn). 

Khaled B. Letaief is with the Department of Electronic and Computer Engineering, The Hong Kong University of Science and Technology, Hong Kong (e-mail: eekhaled@ust.hk).}
}

\maketitle

\begin{abstract}
Advancements in artificial intelligence (AI) and low-earth orbit (LEO) satellites have promoted the application of large remote sensing foundation models for various downstream tasks. However, direct downloading of these models for fine-tuning on the ground is impeded by privacy concerns and limited bandwidth. 
Satellite federated learning (FL) offers a solution by enabling model fine-tuning directly on-board satellites and aggregating model updates without data downloading.
Nevertheless, for large foundation models, the computational capacity of satellites is insufficient to support effective on-board fine-tuning in traditional Satellite FL frameworks.
To address these challenges, we propose a satellite-ground collaborative federated fine-tuning framework. The key of the framework lies in how to reasonably decompose and allocate model components to alleviate insufficient on-board computation capabilities. During fine-tuning, satellites exchange intermediate results with ground stations or other satellites for forward propagation and back propagation, which brings communication challenges due to the special communication topology of space transmission networks, such as intermittent satellite-ground communication, short duration of satellite-ground communication windows, and unstable inter-orbit inter-satellite links (ISLs). To reduce transmission delays, we further introduce tailored communication strategies that integrate both communication and computing resources. Specifically, we propose a parallel intra-orbit communication strategy, a topology-aware satellite-ground communication strategy, and a latency-minimalization inter-orbit communication strategy to reduce space communication costs. Simulation results demonstrate significant reductions in training time to 33\% of on-board training time.
\end{abstract}

\begin{IEEEkeywords}
Satellite federated learning, fine-tuning, foundation models, edge learning, and satellite communications.
\end{IEEEkeywords}

\IEEEpeerreviewmaketitle

\begin{figure*}[t]
    \centering
    \includegraphics[width=1\linewidth]{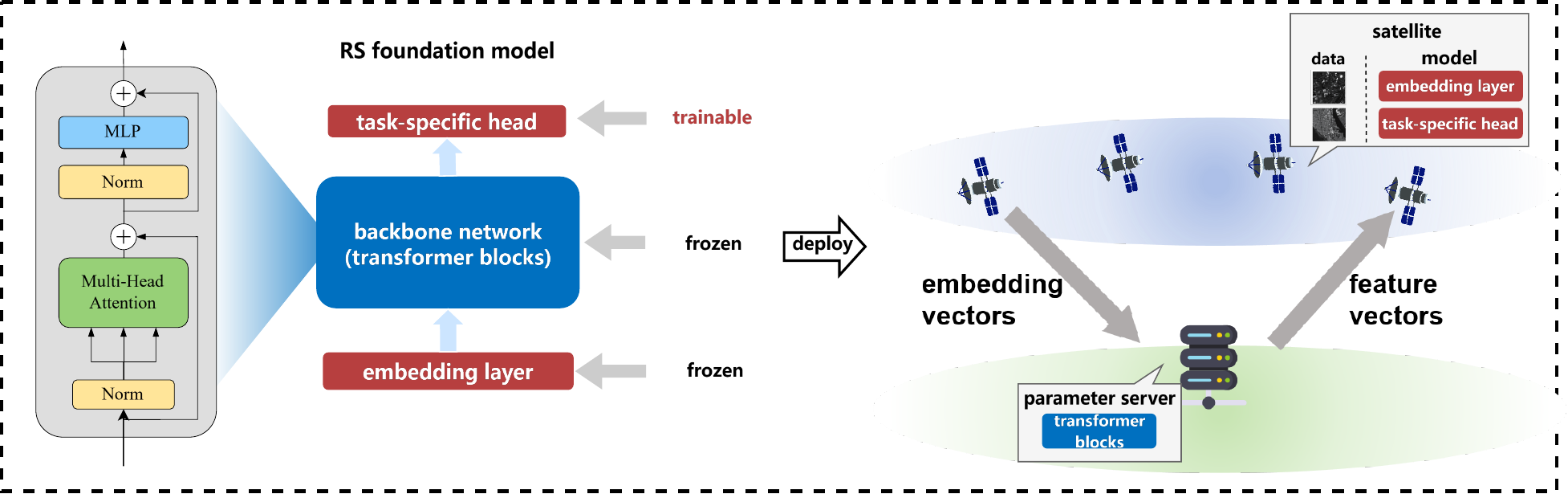}
    \caption{Architecture and deployment of remote sensing (RS) foundation models. The architecture of the RS foundation models is organized into three distinct components: the embedding layer, the backbone network, and the task-specific head. The embedding layer and task-specific head are deployed on satellites, while the backbone network, which handles the primary computation load, is deployed on the ground server.}
    \label{archi_and_deploy}
    \vspace{-0.3cm}
\end{figure*}
\section{Introduction}
The proliferation of low-earth orbit (LEO) satellites, driven by advancements in satellite technology, has established them as indispensable tools for acquiring high-resolution imagery of the Earth's surface. 
These data play a pivotal role in Earth observation applications such as environmental monitoring~\cite{environmentmonitor} and land cover classification~\cite{landuse}. Simultaneously, artificial intelligence (AI) has experienced remarkable growth over recent decades, substantially contributing to the development of remote sensing data interpretation. 
In the remote sensing field, the most commonly used backbone of the learning models can be categorized into convolutional neural networks (CNNs) and Transformers~\cite{RSsurvey}. CNN-based models, with the unique architecture of convolutional layers, can automatically extract hierarchical and discriminative features and preserve spatial information from remote sensing images, such as scene classification \cite{RS2}. Transformer-based models, such as variants of Vision Transformer, capture long-range dependency better than the CNN-based models \cite{ViT}. Transformer-based models are applied in many scenarios in the remote sensing field, including generalist geospatial AI (e.g., Prithvi \cite{Prithvi}), visual tasks (e.g., SatMAE \cite{SatMAE} and SatMAE++ \cite{SatMAEplus}), and spectral data analysis (e.g., SpectralGPT \cite{SpectralGPT}).
These foundation models are pre-trained on large-scale remote sensing datasets and fine-tuned on specific tasks to enhance their performance in various downstream tasks.

Traditional data-driven model training has limitations as it needs to download raw satellite data to terrestrial cloud servers. Firstly, satellite-ground communication faces short communication durations due to satellite movement, which poses a challenge to the real-time transmission of the original data. Secondly, the high resolution of remote sensing images has made the satellite-ground links (SGL) with only a few hundred Mbps become the transmission bottleneck. 
For instance, transmitting a scene at 0.5-meter resolution over a 30-kilometer view width results in approximately 40 GB of data transfer between satellites and ground stations (GSs). Finally, the direct transmission of original data exposes sensitive information, leading to serious privacy and security issues. At the same time, the communication and on-orbit computing technologies of satellite networks are becoming increasingly mature, which lays a foundation for the development of space computing power networks (Space-CPN). 
In Space-CPN, satellites are combined with terrestrial devices to form a network. By performing real-time data processing on-board, Space-CPN can detect and analyze real-time environmental changes, such as forest fires, floods, and climate change. However, fully leveraging the potential of Space-CPN requires an efficient approach to training AI models across distributed satellite nodes while minimizing communication overhead.

A promising solution to this challenge is satellite federated learning (SFL), which enables satellites to collaboratively train AI models without requiring raw data to be transmitted to ground stations~\cite{9749193,edgeAI,shiFEEL}. While SFL has been explored in prior studies, most existing work focuses on small-scale models. Despite the rapid development of Space-CPN, onboard computation capabilities remain significantly constrained compared to terrestrial AI clusters, making the fine-tuning of entire large models a major challenge. Satellite federated learning, as a solution for distributed training, supports collaborative training among satellites~\cite{9749193,edgeAI}.  
To solve this problem, it is a feasible solution to decompose and deploy the computing tasks of fine-tuning large models in Space-CPN and use the distributed computation capabilities for federated fine-tuning~\cite{FedBERT,FFT}. According to the distribution of computation capabilities in Space-CPN, the modules are deployed as follows: the embedding layer is deployed on satellites because satellites are data sources, the backbone network is deployed on the terrestrial server, and the classification layer generating outputs is deployed on satellites, as illustrated in Fig.~\ref{archi_and_deploy}.

In this deployment framework, the primary challenge lies in the communication process. During forward propagation, satellites compute local embedding vectors and perform intra-orbit aggregation. Notably, while raw satellite data can reach gigabyte levels in total size, the transmitted embedding vectors for each sample are only a few megabytes, significantly alleviating communication overhead. Once computed, these embedding vectors are transmitted to the ground for feature vector computation. In this process, satellite-ground communication faces some special challenges compared with transmission on the ground~\cite{FFT}. Firstly, due to the movement of satellites, especially LEO satellites that circle the Earth at high speeds, SGLs are dynamic and unstable. Secondly, it takes much time to access the GS for each satellite. Besides, the insufficient bandwidth of SGLs leads to slower data rates and higher latency. Then, feature vectors are sent back to satellites for output generation, which faces similar questions as embedding vector downloading. Finally, satellites update local models and aggregate a global model through ISLs. The dynamic ISLs make link selection and allocation extremely complex in terms of network management~\cite{5G}.

To address the above-mentioned challenges, in this paper, we propose a satellite-ground collaborative federated fine-tuning framework, which decomposes the modules of remote sensing foundation model based on the distribution of computing power between space and ground networks. During the federated fine-tuning procedures, we address challenges during satellite-ground communication for embedding vector downloading and feature vector uploading, intra-orbit communication for embedding vector aggregation, and inter-orbit communication for global head aggregation. We design the corresponding collaborative transmission method combined with the process of computation tasks. The major contributions of this paper are summarized as follows:
\begin{enumerate}
    \item We propose an efficient satellite-ground collaborative federated fine-tuning framework that strategically decomposites the fine-tuning task for remote sensing foundation models by allocating the modules between GSs and satellites. This framework offers a practical solution for on-board remote sensing foundation model fine-tuning, ensuring data privacy and alleviating satellite-ground communication bottlenecks.
    \item  We introduce customized communication strategies for satellite-ground and inter-satellite transmission according to the inherent characteristics of transmission links (e.g., intermittent SGLs, stable intra-orbit ISLs, and time-varying inter-orbit ISLs). For sporadic satellite-ground communication, we propose a topology-aware communication strategy that allocates transmission tasks for SGLs according to real-time topology and link states. For intra-orbit communication, we propose a parallel communication strategy based on Ring Allreduce for the ring topology of each orbit. For inter-orbit communication, we propose a communication algorithm to minimize latency according to link capacity. These strategies integrate communication with computation processes, taking into account factors such as bandwidth limitations, sparse connection, and dynamic topology, optimize data transmission efficiency, and accelerate model convergence.
    \item We conduct extensive simulations to evaluate the performance of the proposed framework and communication strategies. The simulation results demonstrate that the proposed framework, coupled with the optimized communication strategies, significantly mitigates training latency, reduces transmission overhead, and improves the convergence speed of the model. 
\end{enumerate}

The remainder of the paper is organized as follows. Section~\ref{related work} introduces relevant research on Satellite FL. Section~\ref{system model} formulates satellite-ground collaborative federated learning, introduces communication networks, and outlines the module decomposition and deployment framework. The workflow of the proposed framework is described in Section~\ref{Satellite-ground Collaborative Federated Fine-tuning}. In Section~\ref{optimization}, communication algorithms are designed to optimize transmission procedures. 
Section~\ref{simulation} presents simulation results, followed by the conclusion in Section~\ref{conclusion}.

     \begin{table*}[h]
	\centering
	\caption{Key parameters}
	\footnotesize
	\renewcommand\arraystretch{1.2}
	\begin{tabular}{l|l|l}
		\hline
		  \textbf{data} & \textbf{size} & \textbf{calculcation}
            \\\hline
		image size of 4800 $\times$ 2800 pixels & 1.4Gb & width $\cdot$ height $\cdot$ channels $\cdot$ 8 bits
         \\ embedding vector size of spectralGPT & 20MB & hidden sequence length $\cdot$ hidden dimension $\cdot$ 32 bits
         \\ feature vector size of spectralGPT & 32kb & feature dimension $\cdot$ 32 bits
         \\ Flops of embedding layer & $1.13\times 10^8 $flops &patch number $\cdot$ patch size $\cdot$ embedding dimension
         \\ Flops of transformer block & 5.94$\times 10^8$ flops & patch size$^2\cdot$ hidden dimension + patch size $\cdot$ hidden dimension$^2$
         \\ Flops of backbone network & 7.13$\times 10^9$ flops & the number of transformer blocks $\cdot$ flops of transformer block
         \\ Flops of head & 1.1$\times 10^6$ flops& input dimension $\cdot$ output dimension
         \\ average communication window & 10 minutes & from satellite tool kit (STK)
         \\ average revisit intervals & 3 hours & from STK
         \\\hline
	\end{tabular}
    
    \label{123}
    \end{table*}

\section{Related Works}
\label{related work}
\subsection{Centralized Satellite FL}
In centralized Satellite FL, a coordinator, such as a parameter server (PS), collects model updates from satellites. Satellites perform local training on their datasets and then send the model updates to the central coordinator. Some algorithms employ only SGLs during training, where models are trained on-board and transmitted via SGLs for global model aggregation at the PS. These algorithms are usually designed to address the challenges of time-varying and unstable SGLs. For instance, the algorithm proposed in~\cite{DAFL} adjusted aggregation intervals based on the density of satellite-ground connections to increase the frequency of model updates. \cite{SFL} proposed an over-the-air computation based scheme and optimized both uplink and downlink communications through beamforming. Considering the variations in connection time and connection density among satellites, asynchronous FL is introduced to mitigate the effects of stragglers, allowing satellites to send model updates to the server asynchronously~\cite{9674028,FedSN}. A compensation mechanism proposed in~\cite{FedGSM} addressed gradient staleness by adjusting the updates according to discrepancies.

Satellite FL with SGLs presents significant drawbacks, such as high latency due to satellite-ground communication delays. So some work adopts ISLs to enable model aggregation among satellites, thereby reducing the data transmission via SGLs. Some studies use high-altitude platforms as the coordinator for improved communication efficiency and a more favorable communication environment, such as \cite{A}. FedISL leverages stable intra-orbit ISLs for intra-orbit model aggregation, predicts satellite movement and sends intra-orbit aggregated models through SGLs for global model aggregation~\cite{10409275}. AsyncFLEO utilizes satellites as relays, employs asynchronous FL, and proposes a model propagation algorithm that incorporates satellite grouping and stale update discounts to accelerate convergence, improve model accuracy, and balance the idle time and model staleness~\cite{AsyncFLEO}.
 
\subsection{Decentralized Satellite FL}
To eliminate the reliance on the central coordinator, decentralized Satellite FL is proposed, where satellites exchange model parameters directly with other satellites through ISLs. In this context, the primary focus is on designing inter-orbit transmission schemes aimed at minimizing transmission time or energy consumption. The scheme in \cite{DSFL} aggregates a global model at each training round and strives to minimize energy consumption. Some decentralized Satellite FL incorporates the gossip learning, where satellites communicate with their neighbors for partial model aggregation at each training round~\cite{gossipFL,FedLEO,DFedSat}. \cite{FedLEO} offloads data through ISLs to balance data distribution and optimize delay and accuracy within computation and communication power constraints. DFedSat in \cite{DFedSat} reduces communication costs and improves system robustness by implementing a self-compensation mechanism to address packet loss. \cite{NumberControl} proposed a number control scheme to minimize the convergence time of decentralized Satellite FL through optimization of constellation configurations.

Numerous efforts have been made to accelerate the convergence of Satellite FL by addressing communication challenges such as intermittent satellite-ground communication and dynamic inter-satellite communication. However, most existing works rely on satellites to train models and ignore the challenges of on-board large model training stemming from insufficient computation capacity. To overcome these limitations, we propose a satellite-ground collaborative federated fine-tuning framework to train large models.

\subsection{Existing Communication Strategy}
Prior work on intra-orbit aggregation relies on simple sequential data collection, where a designated satellite aggregates intra-orbit updates before forwarding them. This leads to linear latency growth with the number of satellites, wasting bandwidth due to idle links. Existing strategies for satellite-ground links (SGLs) either ignore inter-satellite cooperation and each satellite transmits local data independently. For example, the method in ~\cite{DAFL} adjusts aggregation intervals based on connection density but does not optimize transmission tasks across multiple SGLs, leading to underutilization of short communication windows. Decentralized SFL methods use inter-orbit ISLs but focus on packet loss compensation rather than latency minimization, such as ~\cite{DFedSat}. Gossip-based inter-orbit transmission requires more rounds to reach global consensus, increasing training epochs.

\section{System model}
\label{system model}

\begin{figure}[t]
    \centering
    \includegraphics[width=0.95\linewidth]{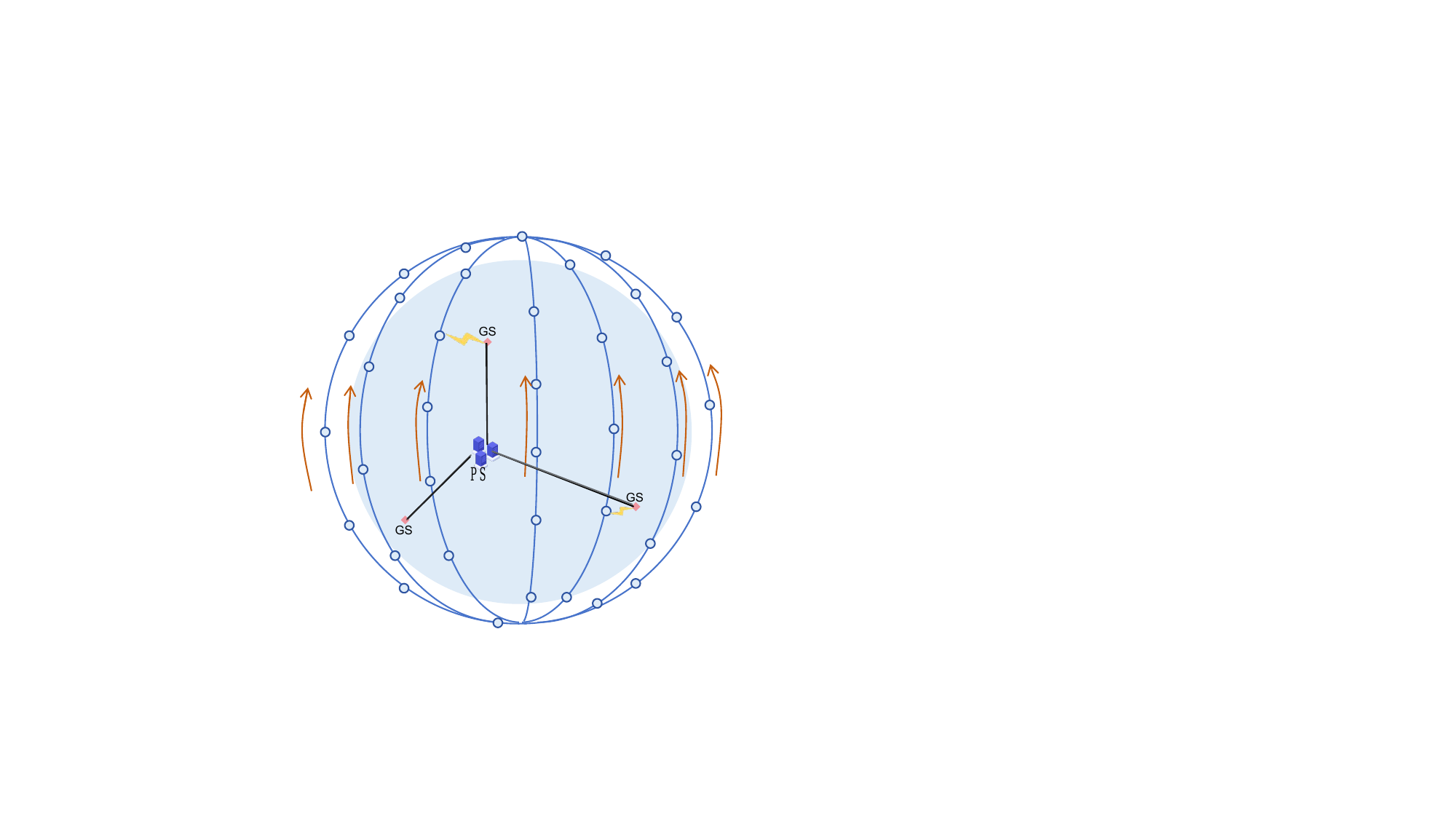}
    \caption{Overview of the walker constellation. Satellites move around the Earth in circular orbits. Multiple GSs are located at different locations.}
    \label{constellation}
    \vspace{-0.3cm}
\end{figure}

Consider a general Walker constellation, which has remarkable advantages in global coverage and is widely utilized in various communication and LEO systems like Starlink, OneWeb, and Kuiper \cite{walker}. The Walker constellation is formulated by the parameters $(M/P/F,H,I)$, where $I$ is the inclination, $M$ is the total number of satellites, $P$ is the number of orbit planes, $F$ is the phase factor of two adjacent orbits, and $H$ is the altitude of orbits. In the LEO satellite constellation, satellites orbit the Earth at an altitude ranging approximately from 500 to 1500 kilometers. Orbit planes and satellites are evenly spaced as depicted in Fig.~\ref{constellation}. Each orbit encompasses $N=M/P$ satellites, and the satellites are indexed by the set $\mathcal{S}=\left\{S_{11},\cdots, S_{1N},\cdots, S_{P1}, \cdots ,S_{PN} \right\}$, where $S_{p,n},p\in\{1,\dots,P\},n\in\{1,\dots,N\}$ denotes the $n$-th satellite in the $p$-th orbit. 

\subsection{Satellite Federated Fine-Tuning}
\label{constellation model}

In Space-CPN, each satellite stores massive remote sensing data within a Walker constellation, which can be harnessed to train AI models with federated fine-tuning for downstream remote sensing tasks, including environmental monitoring and land cover classification. A Satellite FL system comprises a ground PS responsible for coordination and global model aggregation, alongside multiple GSs tasked with facilitating data transmission between satellites and the PS. Each satellite trains its local model on the local dataset without transmitting the raw dataset to the GSs. The goal of Satellite FL is to collaboratively train a global model $\hat{\boldsymbol{w}}$ to minimize the global loss, which is mathematically expressed as:
\begin{align}
\hat{\boldsymbol{w}} = \mathop{\min}_{\boldsymbol{w}} \sum_{i=1}^P
\sum_{j=1}^N \frac{m_{i,j}}{m} L(\boldsymbol{w};D_{i,j}),
\end{align}
where $D_{i,j}$ represents the local dataset on satellite $S_{i,j}$, $m_{i,j} = |D_{i,j}|$ is the size of the local dataset, $m=\sum_{i=1}^P \sum_{j=1}^N m_{i,j}$ is the total number of data samples, and the function $L(\cdot)$ denotes the local loss function dependent on the local dataset $D_{i,j}$ and the local model $\boldsymbol{w}_{i,j}$. Conventional Satellite FL adheres to the subsequent steps:
\subsubsection{Initialization} The ground PS transmits the initialized model $\boldsymbol{w}(0)$ to all satellites, and each satellite initializes its local model by
    \begin{align}
        \boldsymbol{w}_{i,j}(0)\leftarrow \boldsymbol{w}(0).
    \end{align}
\subsubsection{Local Update} At the training round $r$, each satellite updates its local model through the stochastic gradient descent (SGD) algorithm, which is given by
    \begin{align}
        \boldsymbol{w}_{i,j} (r+1) = \boldsymbol{w}_{i,j} (r) - \eta \tilde{\nabla} L(\boldsymbol{w}_{i,j} (r)),
    \end{align}
    where $\eta$ represents the learning rate and $\tilde{\nabla} L(\cdot)$ denotes the stochastic gradient oracle. 
\subsubsection{Global aggregation} The local models from different satellites are first transmitted to GSs and subsequently relayed to the ground PS for global aggregation~\cite{FL}. The global aggregation formula is 
    \begin{align}
    \boldsymbol{w} (r+1) = \sum_{i=1}^P \sum_{j=1}^N \frac{m_{i,j}}{m} \boldsymbol{w}_{i,j}(r+1).
    \end{align}

\subsection{Communication Networks}
\label{communication network}
In Space-CPN, communication serves as the fundamental enabler for data exchange, resource coordination, and overall system operation. Without communication, the individual computing nodes in space would be isolated, so we analyze the characteristics of communication networks.
 
\subsubsection{Ground Communication Network} 
During federated fine-tuning, training information is exchanged between satellites and GSs for global aggregation. As depicted in Fig.~\ref{constellation}, there are $G$ GSs and a PS situated on the ground, and all GSs are connected with the PS. The PS serves as the central coordinator in this system, while the GSs act as relays that facilitate communication between the satellites and the PS. 
The communication links between the GSs and the PS are fixed and characterized by high-speed fabric connectivity, ensuring stable and fast transmission~\cite{groundLinks}. The bandwidth is much larger than the bandwidth of the SGLs, so data transmission between GSs and PS will not become a transmission bottleneck.
To represent the characteristics of the ground communication network, we model it as a stable star network, as illustrated in Fig.~\ref{GS topology}, where the PS is the central node and transmission between the central node and other nodes is fast.

\subsubsection{Satellites Communication Network} 
\begin{figure}[tbp]
    \centering
    \subfloat[][Satellite network: a set of rings.]{\label{satellite topology}\includegraphics[width=.5\linewidth]{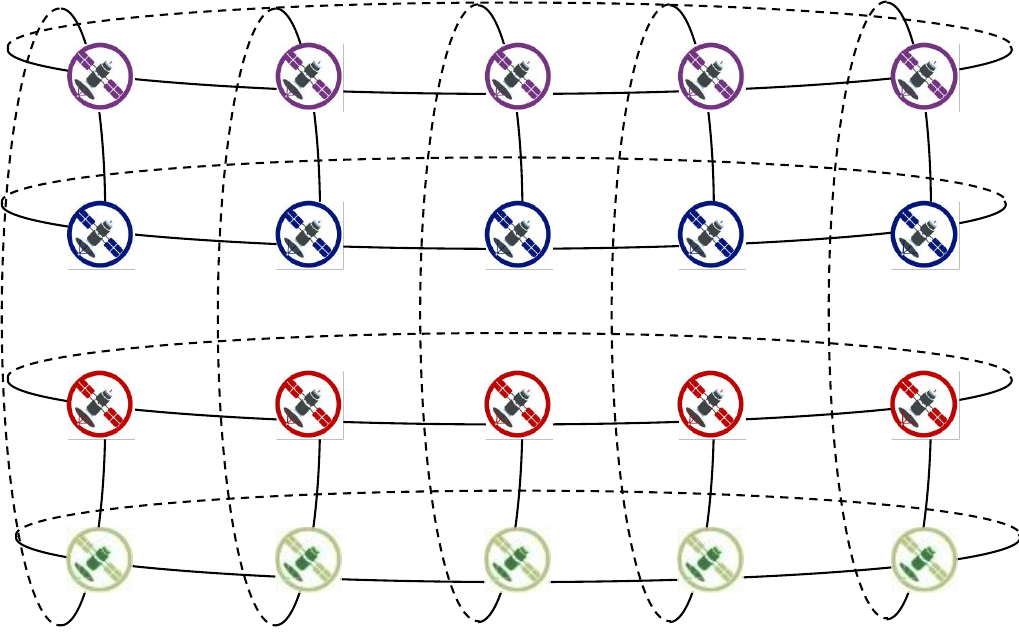}}
    \hspace{1em}
    \subfloat[][Ground network: star topology.]{\label{GS topology}\includegraphics[width=.4\linewidth]{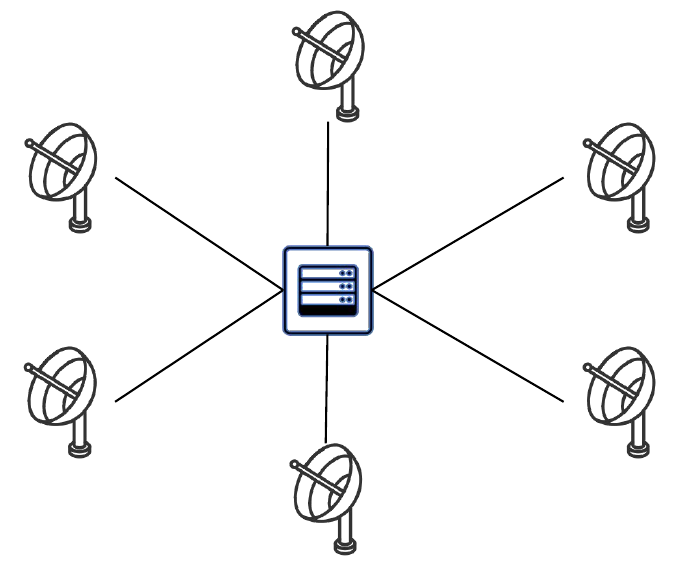}}
    \\
    \subfloat[][Satellite-ground network: time-varying topology.]{\label{satellite ground topology}\includegraphics[width=1\linewidth]{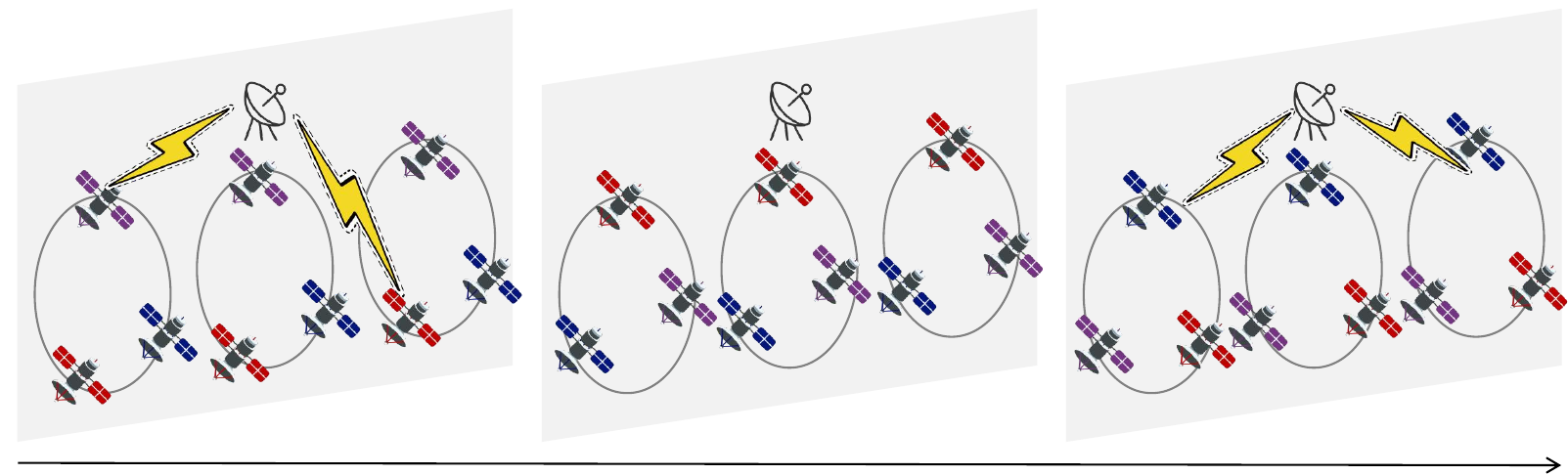}}

    \caption{Satellite-ground communication network topology}
\end{figure}

LEO satellites are capable of communicating with each other through laser ISLs. There exist two distinct types of laser ISLs: intra-orbit ISLs for communication between two adjacent satellites within the same orbit, and inter-orbit ISLs for communication between two satellites in adjacent orbits. 
Each satellite $S_{ij}$ typically establishes four ISLs: two intra-orbit ISLs established with $S_{i,j-1}$ and $S_{i,j+1}$ and two inter-orbit ISLs with $S_{i-1,j_1}$ and $S_{i+1,j_2}$. Laser ISLs are widely utilized as primary inter-satellite communication links by many constellations such as Kuiper, Telesat, and Starlink~\cite{laserISL}. The power received is expressed by
\begin{equation}\label{eq1}
P_{R}=P_{T}G_{T}G_{R}L_{ps}L_{pt},
\end{equation}
where $P_{T}$ is the transmitting power, $G_{R}$ and $G_{T}$ are the receiving and transmitting efficiencies, $L_{ps}$ is the pointing loss, and $L_{pt}$ is the path loss~\cite{ISLs}. The pointing loss due to misalignment from satellite jitter and tracking noise can be expressed as
\begin{equation}  
    L_{ps}=e^{-G_T\theta_T^2-G_R\theta_R^2},
    \end{equation}
where $\theta_T$ and $\theta_R$ are the pointing error angles. The free-space path loss is calculated as
\begin{equation}
\label{path loss}
    L_{pt}=\biggl(\frac{\lambda}{4\pi l}\biggr)^{2},
\end{equation}
where $\lambda$ is the wavelength and $l$ is the distance. The noise is the additive white Gaussian noise with variance $\sigma^2$, so the achievable data rate of laser ISLs is given by
\begin{equation}
    R = B\log_2(1+\mathrm{SNR}) = B\log_2(1+ \frac{P_{R}}{\sigma^2}) ,
\end{equation}
where $B$ is the bandwidth and $SNR$ is the signal-to-noise ratio~\cite{ISL2}. 
Laser ISLs have high data rates and can provide a transmission speed of 100 Gbps~\cite{laserISL1}. The satellite communication network composed of ISLs has the following characteristics:
\begin{itemize}
    \item \textbf{Stable intra-orbit communication.} The intra-orbit ISLs between adjacent satellites ensure stable and consistent connectivity, as these adjacent satellites within the same orbit possess identical velocities and are situated at relatively short distances from one another. All satellites within the same orbit are interconnected end to end, forming a stable ring structure.
    \item \textbf{Unstable inter-orbit communication.} Some satellites in adjacent orbits move in opposite directions in the Walker constellation, as depicted in Fig.~\ref{constellation}. Inter-orbit ISLs between these satellites, known as cross-seam inter-orbit ISLs, face challenges such as the Doppler shift and limited communication windows due to the rapid movement of satellites~\cite{leyva2021inter}.  Consequently, cross-seam ISLs are generally not utilized for inter-satellite communication because of their instability.
\end{itemize}
Leveraging these characteristics, satellites within the same orbit can be modeled as a ring topology. The LEO constellation comprises a collection of such rings, with adjacent rings interconnected by inter-orbit ISLs, excluding cross-seam ISLs, as depicted in Fig.~\ref{satellite topology}.

\subsubsection{Satellite-Ground Communication Network} 
Satellite-ground communication is essential for cooperation during fine-tuning. Satellite-ground channels can be modeled as frequency selective channels with path loss and attenuation~\cite{SGL1}. The definition of free-space path loss remains consistent with Equation~\ref{path loss}. The attenuation due to rain is presented as
\begin{equation}
    L_{rain}=KR_r^\alpha l_r,
\end{equation}
where $K$ and $\alpha$ are the rain attenuation coefficients, $R_r$ is the rain rate, and $l_r$ is the path length in the rain area~\cite{rain}. The received power for each path is expressed as
\begin{equation}
    P_R(i)=\hat{P}_{\mathrm{t}}\left(\frac{\hat{\lambda}}{4\pi}\right)^2\left|\frac{r_ie^{-\Delta\phi_i}}{l_i}\right|^2,
\end{equation}
where $\hat{P}_{\mathrm{t}}$ is the transmit power of SGLs, $\hat{\lambda}$ is the wavelength of satellite-ground signals, $r_i$ is the reflection coefficient, and $\Delta\phi_i$ is the phase difference respect to the direct path. The total multipath received power for $N$ paths is derived as
\begin{equation}
P_{\mathrm{mp}}=10\log_{10}\left[P_{\mathrm{t}}\left(\frac{\lambda}{4\pi}\right)^{2}\left|\frac{1}{l_{1}}+\sum_{i=2}^{N}\frac{r_{i}e^{-j\Delta\phi_{i}}}{l_{i}}\right|^{2}\right].
\end{equation}
The total received power is then:
\begin{equation}
    \hat{P}_r=P_{mp}-L_{pt}-L_{rain}.
\end{equation}
SGLs can provide transmission speeds on the order of Mbps, which is insufficient for satellite data transmission. Due to the movement of satellites, the satellite-ground topology exhibits the following characteristics:

\begin{itemize}
    \item \textbf{Intermittent connectivity.} 
    A satellite becomes visible to a ground station (GS) when it orbits overhead. 
    The period during which the satellite is visible to a GS is termed the communication window. 
    Since satellites are not always within the visible range of GSs due to their movement, it may take hours to move from one visible range to another.  
    Consequently, SGLs experience intermittent connectivity, resulting in sparse satellite-ground links. 
    \item \textbf{Short communication window.} Satellites only communicate with the GS when it is in the visible range. 
    However, due to their rapid movement, LEO satellites quickly pass through this range, leading to short communication windows, typically lasting only a few minutes. 
   
\end{itemize}

When analyzing SGLs, we sample some time points discretely  $\mathcal{T}=\left\{T_0,T_1,T_2,...\right\}$ over a period. A satellite and a GS can communicate between $T_i$ and $T_{i+1}$ if at least one link is feasible at $T_i$. The interval between two adjacent time points is 1 minute. At different time points, the topology of SGLs varies. Thus, we model satellite-ground communication links as time-varying topology, as depicted in Fig.~\ref{satellite ground topology}. For the coordination of multiple ground stations across orbits, a parallel transmission strategy is adopted. Each orbit can independently transmit data through the GSs within its coverage area.

\subsection{Satellite Federated Fine-Tuning for Remote Sensing Foundation Models}
\label{RS model}

Remote sensing foundation models (RS FMs) are specifically designed to analyze RS data from remote sensing devices, such as satellites and drones. These huge FMs are pre-trained and can be fine-tuned to enhance their accuracy and efficiency in specific downstream tasks. Typically, RS FMs consist of the embedding layer, backbone network, and task-specific head, as shown in Fig.~\ref{archi_and_deploy}. Take SpectralGPT as an example, the workflow is as follows~\cite{SpectralGPT}. First, images are split into patches and the embedding layer embeds image patches, adds position embeddings, and outputs embedding vectors. 
Then, the embedding vectors are fed into the backbone network to extract feature vectors. The computation load of the backbone network accounts for the vast majority of the total computation load of the model. 
Finally, for each downstream task, a learnable task-specific head is connected to the model. This head receives feature vectors as inputs and outputs results. During fine-tuning, only head parameters ${\boldsymbol{w}}^H$ are updated, while other parameters remain fixed in each training round $r$.
\begin{align}
\boldsymbol{w}^H_{i,j} (r+1) = \boldsymbol{w}^H_{i,j} (r) - \eta \nabla \operatorname{L}(\boldsymbol{w}^H_{i,j} (r)),
\end{align}
where $\eta$ represents the learning rate, and $\nabla \operatorname{L}()$ denotes the gradient of the loss function.
Since only the head parameters are updated, only head parameters need to be aggregated: 
\begin{align}
\boldsymbol{w}^H (r+1) = \sum_{i=1}^P \sum_{j=1}^N \frac{m_{i,j}}{m} \boldsymbol{w}^H_{i,j}(r+1).
\end{align}
The goal of federated fine-tuning is to find optimal head parameters, similar to Satellite FL. 
\begin{align}
\hat{\boldsymbol{w}}^H = \mathop{\min}_{\boldsymbol{w}^H} \sum_{i=1}^P
\sum_{j=1}^N \frac{m_{i,j}}{m} L(\boldsymbol{w}^H;D_{i,j}).
\end{align}

Our primary objective is to address the computing power constraints of fine-tuning FMs on satellites. To achieve this, we employ a computation task decomposition technique, effectively splitting the FM into distinct components and deploying them based on requirements. 
To protect data privacy, we deploy the embedding layer and the task-specific head on the satellites. 
The backbone network, which bears most of the computational workload (approximately 99\%), is deployed on the ground. This significantly reduces the computational burden on the satellites, allowing them to handle only about $1\text{\textperthousand}$ of the overall computational tasks. 
During the forward propagation process, embedding vectors and feature vectors are exchanged between the satellites and the ground. Notably, the size of these vectors is approximately 20 MB for a sample, a reduction of about a thousand times compared to the raw satellite imagery. This drastic reduction in data size not only enhances communication efficiency but also alleviates bandwidth requirements, making the system more cost-effective and practical for deployment in space-constrained environments.

Based on the scheme of task decomposition, the fine-tuning process can be systematically divided into two distinct stages: feature extraction from satellites to the ground and model update from the ground to satellites.
\subsubsection{Feature Extraction} 
Satellites begin by processing their respective local data through the local embedding layer. The generated embedding vectors are then transmitted from the satellites to the PS located on the ground. The PS further extracts features from embedding vectors by transformer blocks. This ensures that only processed data is sent over the communication link, thereby conserving bandwidth and enhancing security. Overall, this phase focuses on transforming raw satellite imagery into actionable features, with data flowing from satellites to the ground.

\subsubsection{Model Update} 
The feature vectors extracted by the transformer blocks at the ground PS are sent back to the respective satellites. Upon receiving the feature vectors, each satellite processes these vectors and outputs results, computes local loss, and updates local task-specific heads for downstream tasks. 
Finally, all the updated task-specific head parameters from various satellites are aggregated through ISLs. In this phase, data flows from the ground back to the satellites, facilitating a closed-loop learning process. 

Repeat these two stages until the model converges. By decomposing the fine-tuning process into feature extraction and model update stages, we effectively manage computational resources, ensure data privacy, and ensure efficient model synchronization across a distributed satellite network.

\section{Satellite-Ground Collaborative Federated Fine-Tuning}
\label{Satellite-ground Collaborative Federated Fine-tuning}
\begin{figure}[tbp]
    \centering
    \subfloat[Feature extraction.]{\includegraphics[width=0.45\textwidth]{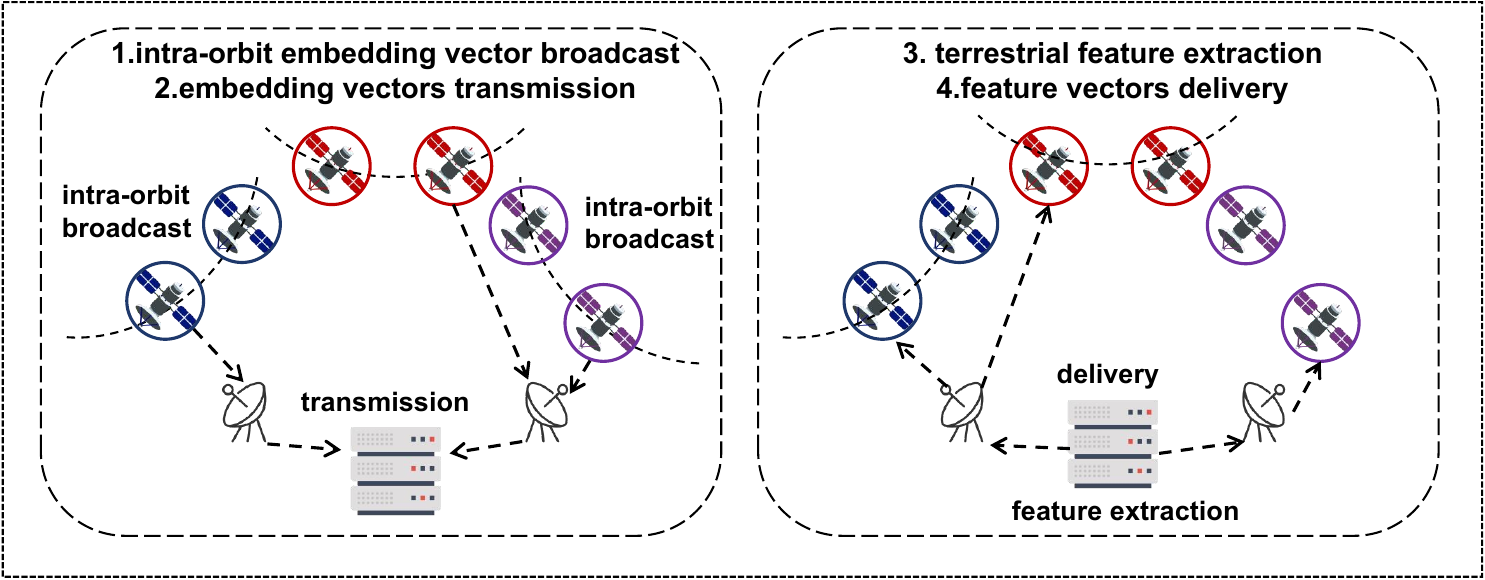}}
    \hfill
    \subfloat[Model update.]{\includegraphics[width=0.45\textwidth]{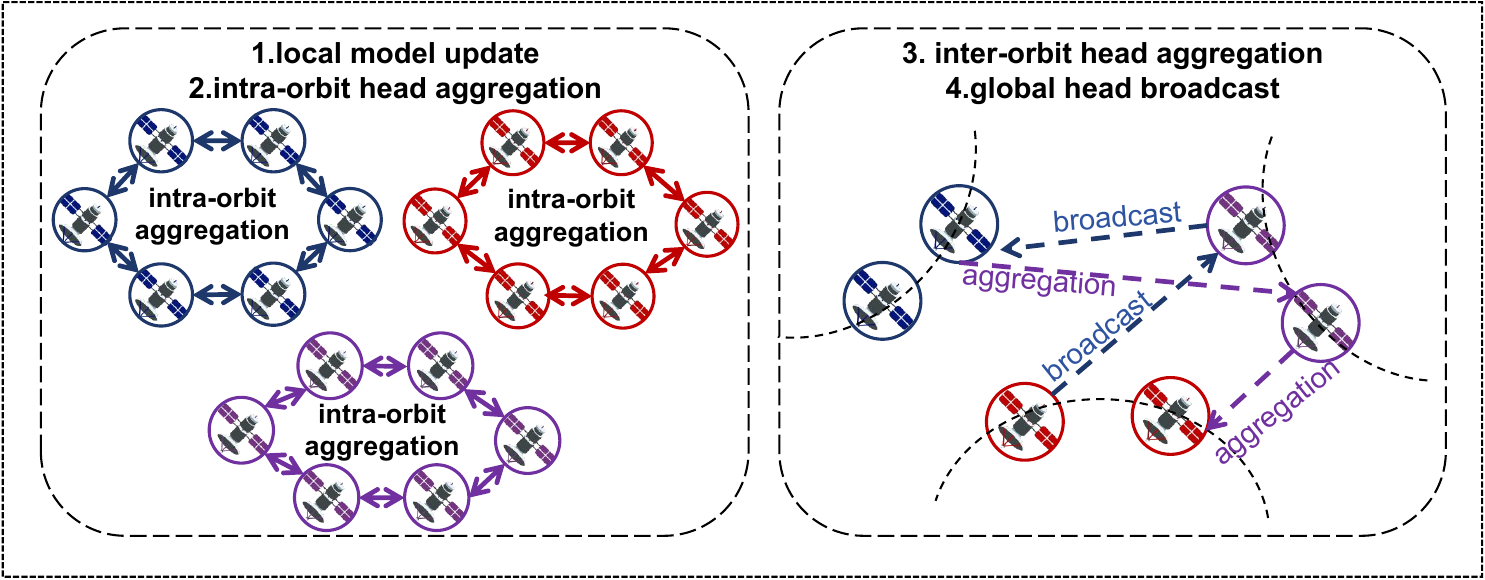}}
    \caption{Workflow of satellite-ground collaborative federated fine-tuning.}
    \label{SG-FFT}
    \vspace{-0.3cm}
\end{figure}

This section provides an overview of the proposed framework, designed to address computational constraints in fine-tuning large foundation models on satellites through a task decomposition mechanism. 
Large FMs require substantial computational resources for fine-tuning, and the limitation in on-board computing power becomes particularly apparent when fine-tuning these models. 
To address these challenges, we propose a task decomposition mechanism to partition the model and allocate different parts of the model to satellites and the PS, respectively. The trainable task-specific head and frozen embedding layer are deployed on satellites, while the frozen backbone network is deployed on the ground. This mechanism significantly reduces the on-board computation load and optimizes the combination of transmission and computation process. We elaborate on the two stages briefly introduced in the section ~\ref{RS model}. Each training round is divided into two phases, feature extraction, and model update, as illustrated in Fig.~\ref{SG-FFT}.

\subsection{Feature Extraction}
\subsubsection{Intra-Orbit Embedding Vector Broadcast}
\begin{figure}[tp]
    \centering
    \includegraphics[width=1\linewidth]{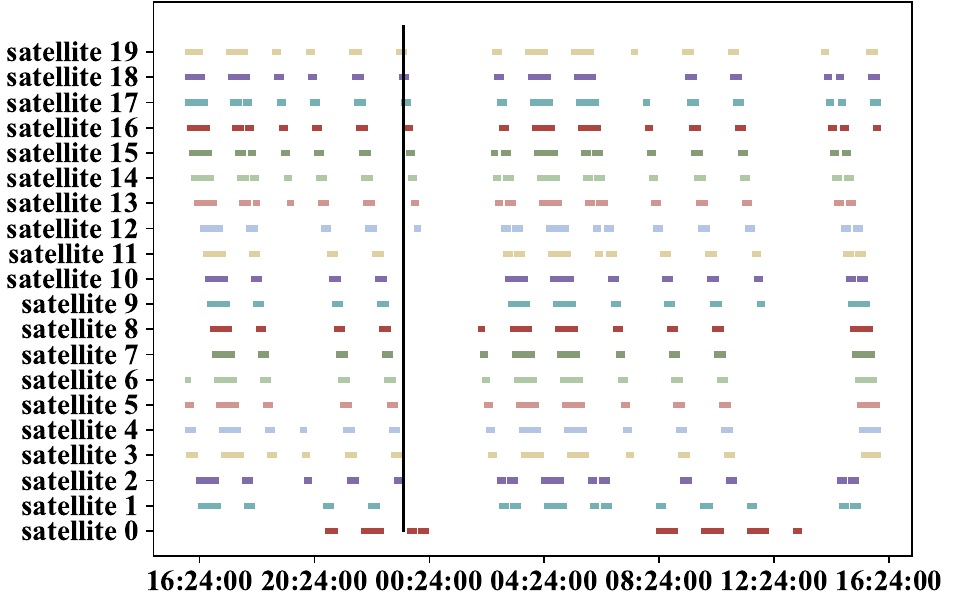}
    \caption{Connection windows between satellites and GSs.}
    \label{connection_window}
    \vspace{-0.5cm}
\end{figure}

First, each satellite $S_{p,n}$ computes embedding vectors $\boldsymbol{E}_{p,n}$ utilizing data stored locally. Then, all computed embedding vectors are broadcast within each orbit so that any satellite connected to a GS can transmit intra-orbit data to the ground for forward propagation. 
We visualize the connection windows between satellites and GSs throughout the day in Fig.~\ref{connection_window} for comparison. Each connection window lasts for approximately $5\sim 10$ minutes. If data transmission starts at 23:30, some satellites establish their first satellite-ground communication around three hours later. So the conventional method, where each satellite only sends its local embedding vectors without coordination, wastes the communication resources of other satellites. In contrast, our proposed method allows other intra-orbit satellites to transmit data immediately without waiting.

Due to the instability of inter-orbit ISLs and the volume of intra-orbit embedding vectors, embedding vectors are not exchanged between different orbits. To optimize this process, we have designed a parallel intra-orbit broadcast strategy, detailed in Section~\ref{intra-orbit strategy}, to make full use of each satellite's communication capabilities. After the broadcast, each satellite stores the concatenated intra-orbit embedding vectors, represented as:

\begin{align}
 \boldsymbol{E}_p= [\boldsymbol{E}_{p,1} , ... , \boldsymbol{E}_{p, N} ].
\end{align}

\subsubsection{Embedding Vectors Transmission}
 Intra-orbit embedding vectors are transmitted from satellites to the PS in each orbit. A key challenge is that satellite-ground communication topology is time-varying, as illustrated in section~\ref{communication network}. To address this, we design a topology-aware satellite-ground transmission strategy in section~\ref{SG strategy}, which allows multiple orbits to independently utilize this strategy for parallel transmission.
The algorithm takes into account the heterogeneity of different SGLs and allocates different transmission tasks to SGLs. Thus, the total transmission time $T_{\text{em}}$ is determined by the maximum transmission time across all orbits: 
\begin{align}
    T_{\text{em}} = \max (T_{\text{em}}^1, \cdots, T_{\text{em}}^P).
\end{align}

\subsubsection{Terrestrial Feature Extraction}
After receiving embedding vectors, the PS feeds them into the backbone network. The backbone network, consisting of dozens of transformer blocks, identifies the most relevant information from the input and outputs feature vectors:
\begin{align}
\boldsymbol{F}_p = \operatorname{f}(\boldsymbol{E}_p) = [\boldsymbol{F}_{p,1} , ... , \boldsymbol{F}_{p, N} ]
\end{align} 
Due to the high computational speed of the PS, the time cost of feature extraction is significantly reduced.

\subsubsection{Feature Vectors Delivery}
Finally, the feature vectors are delivered back to the satellites for the last step of forward propagation. The task-specific head receives these feature vectors and produces the output. This output serves as the basis for further processing in applications. For instance, in classification tasks, the model selects the class with the highest probability as its final prediction. This process also involves satellite-ground communication. Similar to the embedding vector transmission, we use a topology-aware communication strategy for parallel transmission from the PS to satellites in the orbit $p$. Receivers then send the corresponding parts $\boldsymbol{F}_{p,k}$ to other satellites $S_{p,k}$ in the same orbit. 

\subsection{Model Update}
\subsubsection{Local Model Update}
Subsequently, the output is utilized for back propagation in the model. The loss is computed to quantify the difference between the predicted output and the true labels. Once the loss is determined, back propagation is performed to calculate the gradients. These gradients are then used to update the model parameters, typically using an optimization algorithm such as stochastic gradient descent. After this step, the local fine-tuning process for a single satellite is completed. 

\subsubsection{Intra-orbit Head Aggregation}
After local fine-tuning, the updated task-specific heads from all satellites within the same orbit are averaged to ensure model convergence in collaborative learning. Satellites exchange head parameters with each other in a decentralized way without a central terrestrial coordinator. Accounting for the intermittent nature of satellite-ground communication, we perform hierarchical aggregation through ISLs. Head parameters are first aggregated within each orbit
 \begin{align}
  \bar{\boldsymbol{w}}_p^H = \frac{1}{N} \sum_{n=1}^{N} \bar{\boldsymbol{w}}_{pn}^H,
 \end{align}
and then aggregated across different orbits. Due to the more stable communication conditions maintained by satellites within the same orbit, intra-orbit head aggregation employs a parallel communication strategy similar to that used in the embedding vector broadcast. This approach ensures efficient transmission.

\subsubsection{Inter-orbit Head Aggregation}
Intra-orbit heads from all orbits are aggregated to form a global head.
\begin{align}
\bar{\boldsymbol{w}}^H = \frac{1}{P} \sum_{p=1}^{P} \bar{\boldsymbol{w}}_p^H.
\end{align}
Head parameters are transmitted from orbit 1 to orbit $P$ through inter-orbit ISLs for global aggregation. To ensure efficient convergence and communication, the inter-orbit ISLs are carefully selected for transmission between two adjacent orbits. A path selection algorithm is proposed in section~\ref{inter orbit communication}, which can adapt to dynamic inter-orbit ISLs and automatically select the optimal path to minimize transmission latency.

\subsubsection{Global Head Broadcast}
Finally, the global model is transmitted back from orbit $P$ to orbit $1$ via inter-orbit ISLs. Within each orbit, the satellite receiving the global head then disseminates it to other satellites for the next round of training, where satellites continue to train using their local data. This step ensures that every satellite has access to the latest version of the global model. Notably, this can be efficiently achieved by simply reversing the direction of transmission during inter-orbit head aggregation.

In conclusion, each training round of the fine-tuning process comprises two primary phases: feature extraction and model update. These stages are repeated until the model converges.
The entire fine-tuning procedure is delineated in Algorithm \ref{satellite FedFT procedure}. During the training procedure, only the heads and embedding layers are computed on the satellites. By selectively computing only the most relevant or impactful parts of the model, this approach significantly reduces the computational load and offers a promising solution to address the computational challenges, particularly beneficial for resource-intensive model training.

\begin{algorithm}[h]\small
    \SetAlgoLined
	\caption{Satellite-Ground Collaborative Federated Fine-Tuning Algorithm}
        \label{satellite FedFT procedure}
        
        PS initializes and sends model parameters  $\textbf{w}(0)$\ to satellites\;
	\For{synchronous round r=0,...,R }{
            \tcp{feature extraction}
            \For{orbit $p \in \mathcal{P}$ \textbf{parallel}}{
                \For{satellite $n \in\mathcal{K}$ \textbf{parallel}}{
                    Compute local embedding vectors $\textbf{E}_{p,k}(r)$\;
                    
                }
                A satellite gathers and broadcasts concatenation of intra-orbit embedding vectors $\textbf{E}_p(r)$\ \;
                send intra-orbit aggregated embedding vectors $\textbf{E}_p(r)$\ to PS\;
            }
            PS computes and sends feature vectors $\textbf{F}_p(r)$\ to orbit $p$\;
            \tcp{model update}
            \For{orbit $p \in \mathcal{P}$ \textbf{parallel}}{
                 \For{satellite $n \in\mathcal{K}$ \textbf{parallel}}{
            update local model $\textbf{w}^{p,k}_H(r)=\textbf{w}^{p,k}_H(r)-\eta \nabla L(\textbf{w}^{p,k}(r),D^{p,k})$\;
                }
               Aggregate intra-orbit head
                $\bar{\textbf{w}}^p_H(r)=\frac{1}{K} \sum_{i=1}^{N}\textbf{w}^{p,i}_H(r)$\;
            }
            Aggregate the global head $\bar{\textbf{w}}_H(r+1)=\frac{1}{N} \sum_{j=1}^{P}\bar{\textbf{w}}^j_H(r)$\;
            Sends the global head to all orbits\;
	}
	\KwOut{$\bar{\textbf{w}}^H(R)$}
\end{algorithm}

\section{Space Communication Strategies}
\label{optimization}
\subsection{Unified Optimization Framework}
During fine-tuning, a substantial volume of crucial data, such as embedding vectors, feature vectors, and model parameters, is transmitted through ISLs and SGLs. Although the size of these data is significantly smaller than that of the original satellite data, data transmission still accounts for a considerable portion of the overall latency. To achieve efficient transmission, communication is optimized to address various challenges. For satellite-ground communication, SGLs are sporadic and intermittent, with short communication windows.
Satellites usually need to wait for a long time to establish SGLs. For inter-satellite communication, links are dynamic. Therefore, in this section, we present the customized communication strategies for the satellite-ground, intra-orbit, and inter-orbit communication to accelerate convergence. The communication optimization is closely integrated with the computation process.

The proposed communication strategies (Ring AllReduce, topology-aware scheduling, latency-minimal inter-orbit routing) can be unified under a joint communication-computation optimization framework with the objective of minimizing total training latency, defined as:

    \begin{align*}
     \min_{\mathbb{P}, \mathbf{x}, \mathbf{y}} \quad & T_{\text{total}} = T_{\text{comp}} + T_{\text{comm}} \\
    \text{s.t.} \quad & \sum_{n=1}^N x_n = 1 \\
    & x_n = \frac{1}{N}\\
    & \sum \mathbf{y}_l \leq C_l \\
    &  \mathcal{O}(\mathbb{P}) = \{0,1,...,P\} \\
    & T_{\mathbb{P}} = \sum (T_p + T_q)
    \end{align*}

where \(T_{\text{comp}}\) is the total computation time (sum of on-board embedding computation, ground backbone network computation, and local head updates); \(T_{\text{comm}}\) is the total communication time, which depends on fraction of data transmitted per satellite in Ring AllReduce (\(\mathbf{x}\), satellite-ground transmission scheduling \(\mathbf{y}\) and inter-orbit path selection \(\mathbb{P}\). Constraint 1 ensures that Ring AllReduce perform full data aggregation and uniform splitting to achieve minimal latency. Topology-aware scheduling must satisfy flow conservation for link capacity $C_l$ as contraint 2. Constraints 3 and 4 ensure inter-orbit path \(\mathbb{P}\) must cover all orbits (\(\mathcal{O}(\mathbb{P}) = \{0,1,...,P\}\)) for synchronous aggregation and minimize transmission time\(T_{\mathbb{P}} = \sum (T_p + T_q)\), where \(T_p\) and \(T_q\) are propagation and transmission delays. This framework unifies the three strategies under a single latency-minimization objective with constraints tailored to each communication scenario.

\subsection{Intra-Orbit Communication for Partial Head Aggregation}
\label{intra-orbit strategy}
In certain stages, such as intra-orbit embedding vector broadcast and intra-orbit head aggregation, data is transmitted through intra-orbit ISLs. These processes can be abstracted into a unified communication model, where all communication clients establish a stable ring topology. Each client possesses local data that needs to be aggregated or concatenated with the data from other clients before being broadcast to all clients within the ring. If data is transmitted sequentially in each orbit, the transmission time increases as the number of satellites grows. In mega-constellations, the transmission delay can become significant.

To address this issue, we leverage the Ring Allreduce algorithm to accelerate intra-orbit communication. The Ring Allreduce algorithm is a distributed method designed to efficiently aggregate data across multiple clients in a ring topology. In essence, Allreduce means that each client contributes its data, and subsequently, every client receives the result of the reduction. The Ring Allreduce algorithm is particularly communication-efficient for achieving Allreduce in a ring topology. Inspired by the Ring Allreduce algorithm, we divide intra-orbit model transmission into two phases: a scatter-reduce phase and a gather phase.

\begin{figure}[t]
    \centering
    \subfloat[Scatter-reduce phase.]{\includegraphics[width=0.45\textwidth]{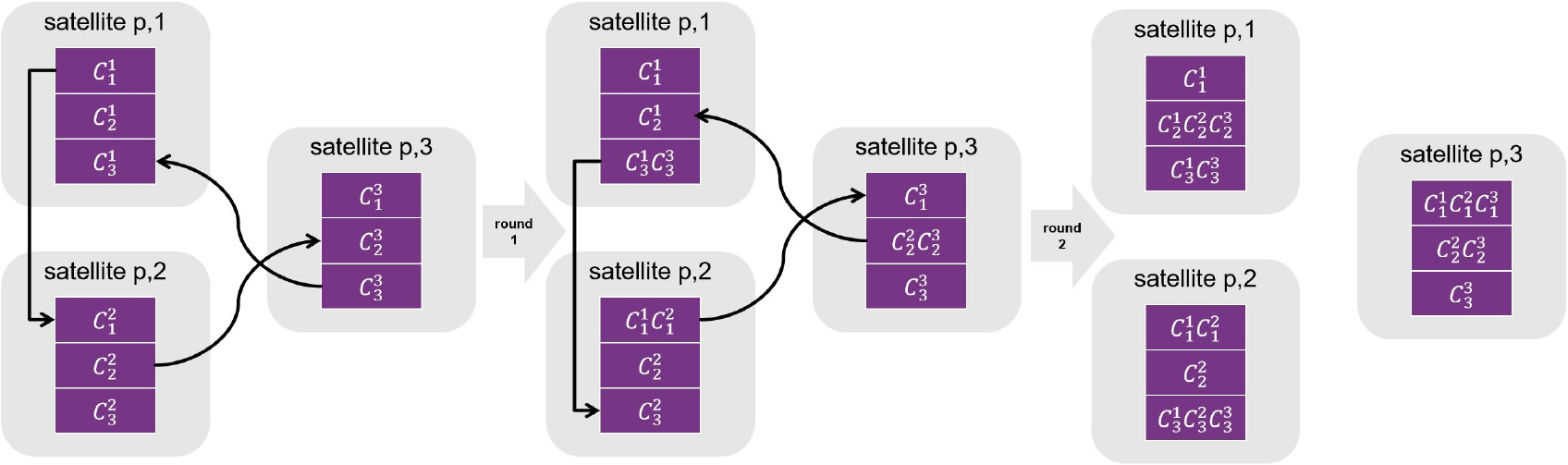}\label{scatter-reduce}}
    \hfill
    \subfloat[Gather phase.]{\includegraphics[width=0.45\textwidth]{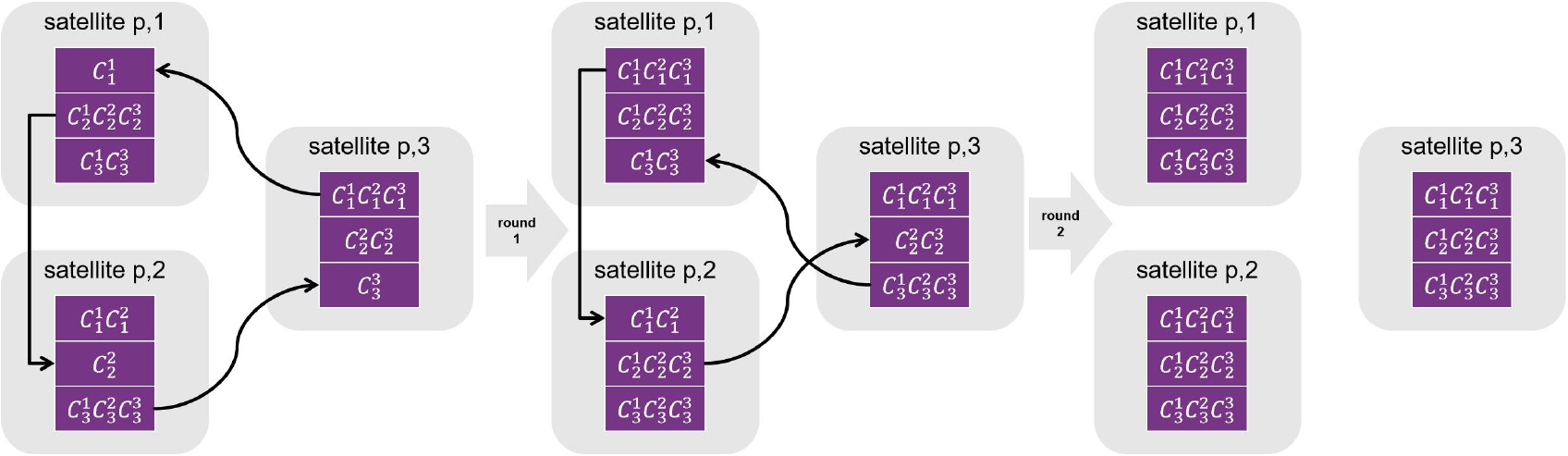}\label{gather}}
    \caption{Ring Allreduce based intra-orbit communication. Take N=3 as an example. Satellites $\left\{ S_{p,1}, S_{p,2}, S_{p,3} \right\}$ travel in orbit $p$. In the scatter-reduce phase, data on satellite $S_{p,i}$ is split into $3$ chunks, i.e., $C_1^i, C_2^i$ and $C_3^i$. At communication round $1$, $S_{p,1}, S_{p,2}, S_{p,3}$ transmit $C_1^1, C_2^2, C_3^3$ respectively. So each satellite only transmits $\frac{1}{3}$ data $2$ times in the scatter-reduce phase and gather phase. The total time is $2 \times 2 \times \frac{1}{3} \times T$, where $T$ is the time to transmit all data on a satellite.}
    \vspace{-0.3cm}
\end{figure}
\subsubsection{Scatter-Reduce Phase}
Each client $S_{p,n}$ splits its local data into $N$ equal-sized chunks, indexed by $\mathcal{C} = \left\{C_1, C_2,...,C_N \right\}$.
During the communication round $r$, each client exchanges its local chunk $C_i$ with its neighbors $S_{p,n+1}$ and $S_{p,n-1}$, where $i=(n-r)\%N+1$. Each client sends a chunk to the next client in the ring and receives a chunk from the previous client. After $N-1$ rounds, each client holds the aggregation result of one chunk from all the clients, as illustrated in Fig.~\ref{scatter-reduce}.
\subsubsection{Gather Phase}
In this phase, each client sends the aggregated chunk to the next client and receives another aggregated chunk from its previous neighbor. After $N-1$ rounds, each aggregated chunk has traveled to all clients, allowing each client to have the complete data, as depicted in Fig.~\ref{gather}.

In the first and second phases of Ring Allreduce, each satellite receives data $N-1$ times, with each transmission involving $\frac{\operatorname{size}(data)}{N}$ bits of data, where $\operatorname{size}(data)$ denotes the total number of bits for size. Thus, when data rate is $r$, the total time is given by: 
\begin{align}
T_{\text{intra}} = 2\cdot(N-1)\cdot\frac{\operatorname{size}(\text{head})}{N\cdot r} < 2\cdot \operatorname{size}(\text{head}).
\end{align}
Ring AllReduce has a constant communication complexity, meaning the communication cost is independent of the number of satellites, effectively mitigating the latency issue. Since each satellite sends and receives data simultaneously in each round, Ring AllReduce can make better use of parallelism and reduce idle time. It is evident that the time consumed by the Ring Allreduce-based transmission scheme is independent of the number of satellites. Therefore, the algorithm scales well with the number of satellites, making it highly suitable for large-scale constellations.

Based on the first phase of the ring all-reduce, we further optimize the intra-orbit orbit of embedding vectors, though they are not reduced but concatenated. In each round, all satellites transmit local vectors and receive vectors in a certain direction. Thus, the transmission time is reduced to:
\begin{align}T_{\text{bc}} = \frac{(N-1) \cdot c_{\text{em}}}{r}\end{align},
where $c_{em}$ is the size of embedding vectors.

\subsection{Satellite-Ground Communication for Embedding Vector Transmission}
\label{SG strategy}
SGLs are sparse, time-varying, and bandwidth-limited. Each satellite communicates with GSs several times daily, and communication windows are short. There is a long time interval between the establishment of SGLs for a satellite. These factors significantly influence the transmission time. We have devised the training steps where data is first broadcast within each orbit and then transmitted through SGLs to fully utilize each transmission opportunity and minimize idle time. However, we did not previously explain the strategy for satellite-ground communication. This section focuses on efficient satellite-ground transmission in each time slot. We have developed a topology-aware communication strategy aimed at maximizing data transmission within each communication window to reduce waiting time. To optimize the utilization of communication resources and shorten the transmission time, we adopt a multi-path parallel transmission scheme that leverages all feasible SGLs.

To effectively organize transmission tasks across heterogeneous SGLs, we model the satellite-ground communication network as a static flow network at each time point. This network is represented as a directed acyclic graph, wherein each edge possesses a capacity constraint, ensuring that the data flow does not exceed this limit. Data flow originates from the source vertex and is sent to the sink vertex. In this context, the source vertex represents the orbits, while the sink vertex corresponds to the PS. 
Data traverses from each orbit to the PS via intermediate satellites and GSs, which are depicted as vertices within the graph. The objective of satellite-ground transmission is to maximize the data flow through the network.

To achieve this objective, we first define the capacities of the edges. For any link $l$, the capacity of its corresponding edge $e_l$ is $\frac{{T}\cdot{r_{l}}}{\alpha_p}$, where $\alpha_p$ denotes the total number of bits of data in orbit $p$, $r_l$ denotes the data rate of link $l$, and $T$ denotes the interval between two time points. Consequently, the capacity of each edge reflects the maximum proportion of data that can be transmitted over the corresponding link during this interval. Subsequently, we apply a max flow algorithm to this flow network to maximize the amount of data transferred from the source to the sink. Each satellite transmits the data according to the algorithm's results until all data is transmitted to the PS. This approach fully leverages the satellite transmission opportunities and ensures the rapid transmission of data without exceeding the transmission capacity of the links. 

\subsection{Inter-Orbit Communications for Global Head Aggregation}
\label{inter orbit communication}
Satellite-ground communication suffers from intermittent SGLs, and the head aggregation does not necessitate the involvement of GSs. Therefore, we propose an aggregation scheme tailored for dynamic satellite networks, wherein satellites aggregate the global head via ISLs to avoid reliance on GSs and unreliable SGLs. The primary challenge for inter-orbit communication lies in the instability and time variability of inter-orbit ISLs. There are different aggregation algorithms designed according to the network topology of the clients, such as the fully connected topology, the ring topology, and so on. However, these designs prove ineffective in LEO satellite networks due to their neglect of satellite network characteristics: the stable ring topology of intra-orbit ISLs and the unstable topology of inter-orbit ISLs.

To propose an aggregation algorithm, it is imperative first to elucidate the network topology. 
We model the LEO satellite constellation as a network where vertices represent satellites and edges denote ISLs. Given the stability of the intra-orbit ISLs, satellites within the same orbit can be considered as a tightly connected cluster and hold the same data because intra-orbit aggregation is done before inter-orbit aggregation. We enumerate the clusters from $1$ to $P$. There are a few links between cluster $i$ and cluster $i+1$, $\forall i \in \left\{1...P-1\right\}$. Cross-seam ISLs are not considered in the system, so there are no links between cluster $1$ and cluster $P$. Our goal is to select a path covering all orbits to achieve fast and stable transmission from cluster $1$ to cluster $P$. The definition of path $P_{s\rightarrow t}$ is the trajectory along which data is transmitted from $s$ to $t$ through certain relay satellites. For example, $\mathbbm{P}_{S_{\text{ab}} \rightarrow S_{\text{cd}}}$ signifies the path from satellite $S_{\text{ab}}$ to satellite $S_{\text{cd}}$ and can be expressed as $\left\{ S_{\text{ab}},S_{x_1y_1},S_{x_2y_2},...,S_{\text{cd}} \right\}$, where $S_{x_iy_i}$ is i-th relay satellite, $S_{\text{ab}}$ is data source and $S_{\text{cd}}$ is the destination. $\mathbbm{P}_{O_{\text{a}} \rightarrow O_{\text{c}}}$ denotes the path from any satellite in orbit $a$ to any satellite in orbit $c$.
To complete global aggregation, the selected path $\mathbbm{P}$ must satisfy:
\begin{align}
\mathcal{O}(\mathbbm{P})=\left\{0,1,\cdots,P\right\},
\label{st1}
\end{align}
where $\mathcal{O}()$ computes the set of covered orbits in path $\mathbbm{P}$. This constraint guarantees that the path $\mathbbm{P}$ can interconnect all orbits.
The global model broadcast can be easily implemented by reversing the direction of transmission in the algorithm. 

To accelerate transmission, path selection is crucial. The delay of data transmission comprises two types: transmission delay and propagation delay. Propagation delay is primarily influenced by the physical medium through which the signal travels and the distance between two satellites. Longer distances result in longer propagation delays. The distance between two satellites $S_{ab}$ and $S_{cd}$ is given by:
\begin{align}\nonumber
    ||S_{ab}S_{cd}||= &((h_a+R_E)^2+(h_c+R_E)^2\\\nonumber
    &-2(h_{a}+R_E)(h_{c}+R_E)
    \\ &\times(\cos{\theta_{ab}}\cos{\theta_{cd}}+\cos{(\epsilon_{a}-\epsilon_{c})}\sin{\theta_{ab}}\sin{\theta_{cd}})),
\end{align}
where $R_E$ is the radius of the Earth. $h_a$, $h_c$, $\epsilon_a$ and $\epsilon_c$ indicate the height of orbit $a$, the height of orbit $c$, the longitude of orbit $a$, and the longitude of orbit $c$, respectively. The latitudes of satellites $S_{a,b}$ and $S_{c,d}$ are denoted as $\theta_{a,b}$ and $\theta_{c,d}$. The propagation delay between satellites $S_{a,b}$ and $S_{c,d}$ is given as:
\begin{align}
    T_p(S_{ab},S_{cd}) = \frac {||S_{ab}S_{cd}||} {c},
\end{align}
where $||S_{ab}S_{cd}||$ is the distance and $c$ is the propagation speed of the signal in the transmission medium. Typically, when laser propagates in a vacuum, $c$ is approximately $3\times10^8$ meters per second. 

Transmission delay is influenced by two factors: the size of the data and the data rate of links between the sender and the receiver. So the transmission delay for transmitting $Z$ bits between satellites $S_{ab}$ and $S_{cd}$ is
\begin{align}
    T_q(S_{ab},S_{cd}) = \frac{Z}{r_{L_{(S_{ab},S_{cd})}}}.
\end{align}
 Data transmission between two satellites $S_{ab}$ and $S_{cd}$ costs time:
\begin{align}
    T(S_{ab},S_{cd}) = T_p(S_{ab},S_{cd}) + T_q(S_{ab},S_{cd}).
\end{align}

\begin{figure}[t]
    \centering
    \includegraphics[width=0.85\linewidth]{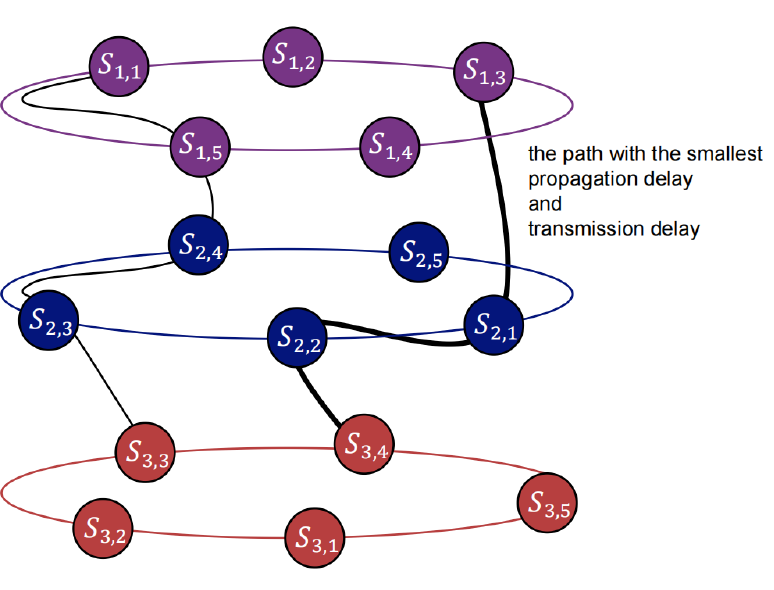}
    \caption{Shortest path problem. The distance between two satellites is set to the sum of propagation delay and transmission delay. So the shortest path is the path with the minimum delay.}
    \label{shortest path}
    \vspace{-0.3cm}
\end{figure}

The total communication delay for global head aggregation is the time it takes for parameters to transmit from orbit $1$ to orbit $P$ on selected path $\mathbbm{P}_{O_{1}\rightarrow O_{P}}$, expressed as the sum of the transmission delay on each ISL belonging to the path:
\begin{align}
\begin{aligned}
T_{\mathbbm{P}_{O_{1}\rightarrow O_{P}}} =  &T(S_{ab},S_{x_1y_1})+T(S_{x_1y_1},S_{x_2y_2})+\cdots
\\
&+T(S_{x_ky_k},S_{cd}).
\end{aligned}
\end{align}
To achieve communication-efficient head aggregation in the LEO satellite network, the problem can be formulated as minimizing the total communication delay as follows.
\begin{align}
\mathscr{P}:~\operatorname{minimize}~T_{\mathbbm{P}}
\end{align}
The solution needs to satisfy the constraint in (\ref{st1}).
To find a path with minimum delay, the distance between satellites $S_{a,b}$ and $S_{c,d}$ is set to $\frac{||S_{a,b}S_{c,d}||}{r_{L_{(S_{a,b}, S_{c,d})}}}$, which is positively correlated with the delay $ T(S_{ab}, S_{cd})$. The minimum delay problem is modeled as a shortest path problem. We run the Floyd-Warshall algorithm on the network to find the shortest path between each pair of satellites. Subsequently, the shortest path between orbit $1$ and orbit $P$ is selected, ensuring that transmitting data on this path incurs the shortest time. 

\subsection{Complexity Analysis}
Two algorithms are mainly used in this paper: the maximum flow algorithm and the shortest path algorithm. Specifically, the time complexity of the Edmonds-Karp algorithm for maximum flow is $O(V\cdot E^2)$, where E denotes the number of edges and V represents the number of nodes. In the context of satellite-ground transmission, V is the sum of the number of ground stations and the number of satellites connected to these ground stations, and E is the number of satellite-ground links. For the Floyd-Warshall algorithm for shortest paths, its complexity is $O(V^3)$. Routing tasks, such as determining the shortest path for inter-orbit communication (e.g., using the Floyd-Warshall algorithm to minimize latency), can be performed on the ground server, similar to routing algorithms in ~\cite{routing1, ROUTING2}, which have significantly stronger computational power compared to on-board satellite systems. These routing results are then transmitted back to the satellites along with feature vectors or embedding vectors. Centralizing complex routing on the ground with robust computing capabilities makes the overall system feasible, given the complexity of $O(V\cdot E^2)$ and $O(V^3)$.

\begin{figure*}[t]
    \centering
    \subfloat[Test accuracy on EuroSAT]{
        \includegraphics[width=0.235\linewidth]{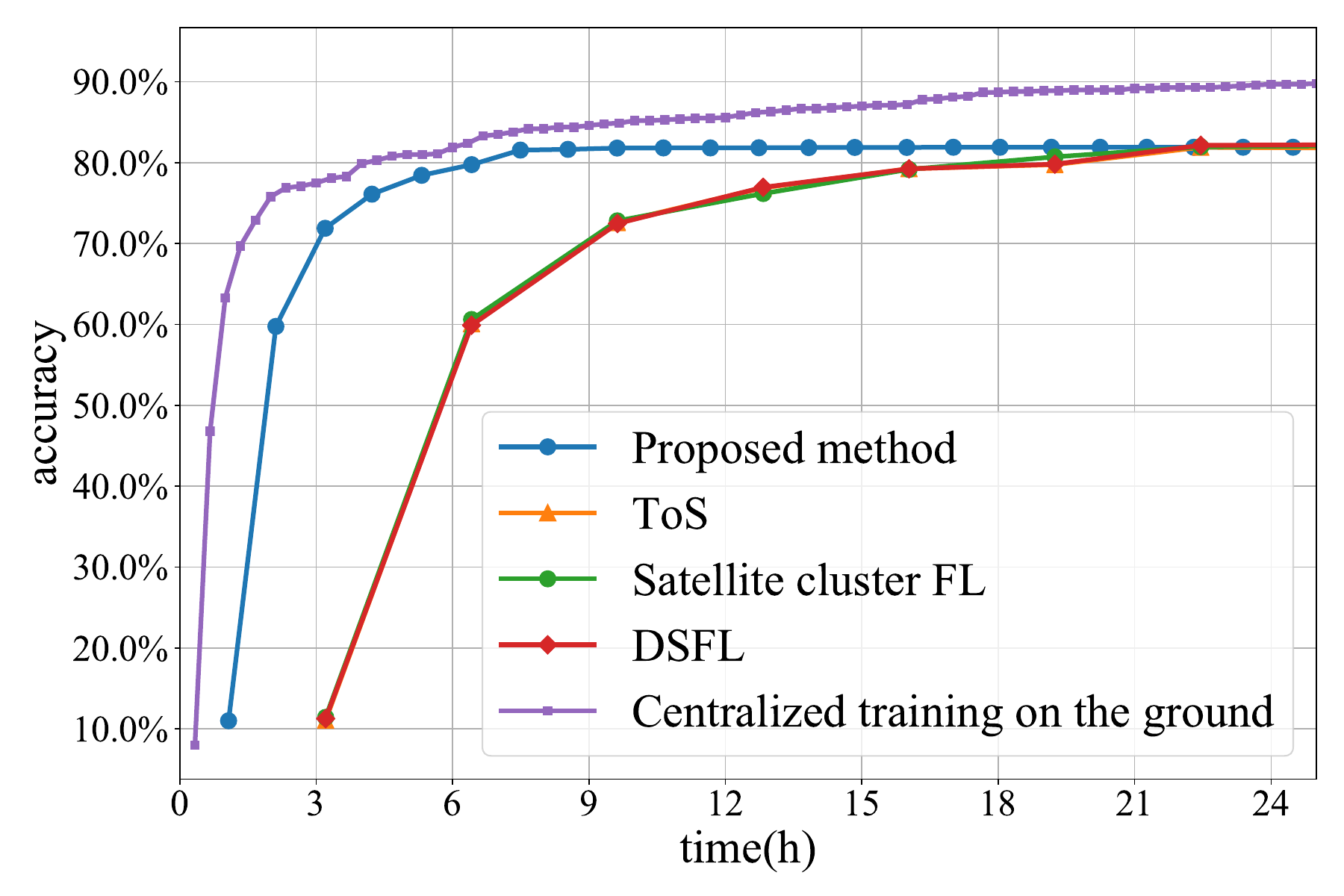}
        }
    \subfloat[Training loss on EuroSAT]{
        \includegraphics[width=0.235\linewidth]{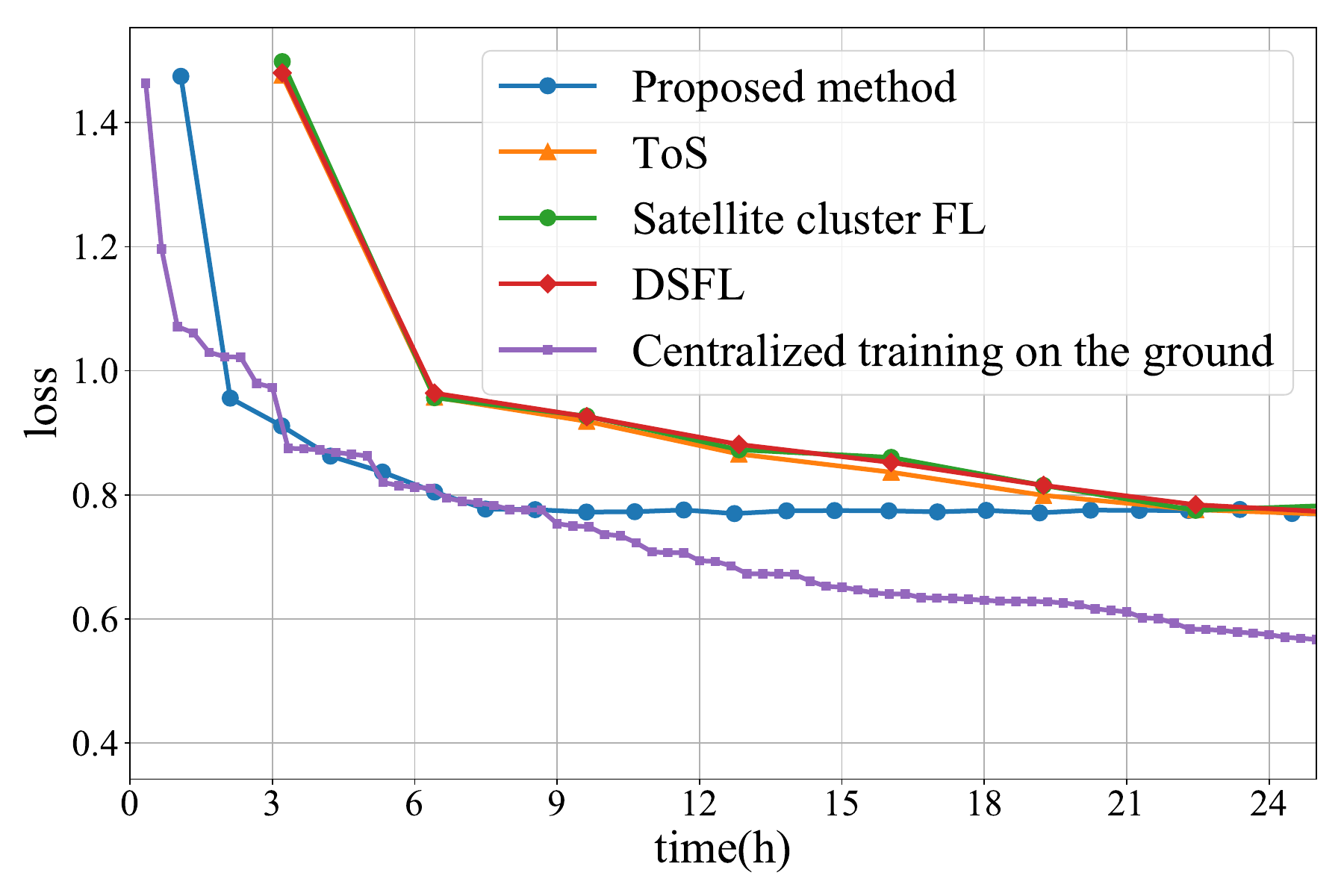}
        }
    \subfloat[Test accuracy on SegMunich]{
        \includegraphics[width=0.235\linewidth]{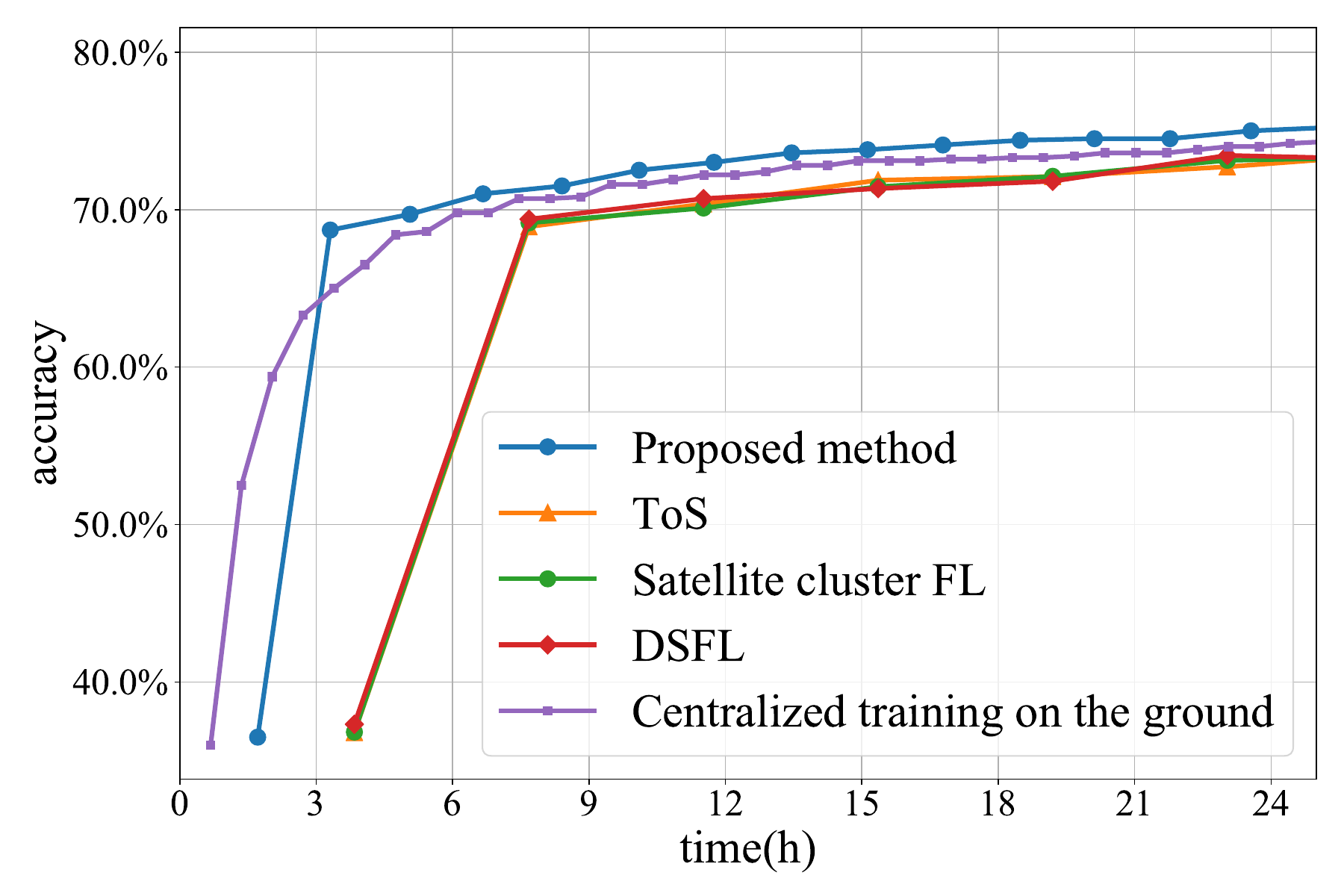}
        }
    \subfloat[Training loss on SegMunich]{
        \includegraphics[width=0.235\linewidth]{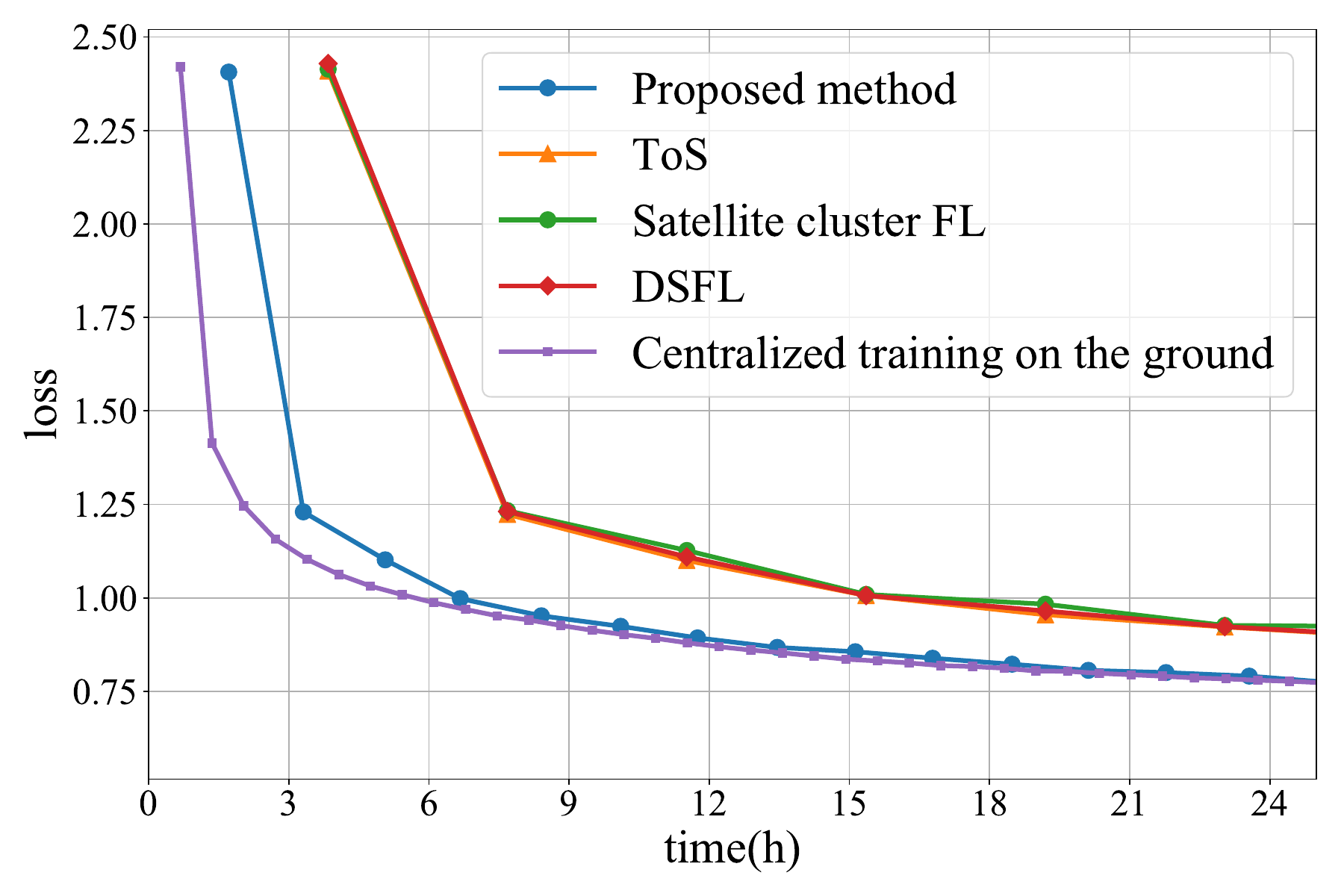}
        }
    \caption{Learning performance versus training time.}
    \label{experiment1}
    \vspace{-0.5cm}
\end{figure*}

\section{Simulations}
\label{simulation}
To evaluate the performance of the proposed framework, we finetune SpectralGPT for different downstream tasks in this section.
\subsection{Simulation Setup}
\subsubsection{LEO Constellation Configurations} We conduct simulations under $80/4/1$, $160/4/1$ and $200/10/1$ Walker constellation setup. $80/4/1$ constellation consists of 4 orbits, with 20 satellites evenly distributed in each orbit. $160/4/1$ constellation is composed of 4 orbits with 40 satellites in each orbit. $200/10/1$ constellation contains 10 orbits, with 20 satellites in each orbit.  All satellites operate at an altitude of 590 km. The parameters for the transmitters and receivers, as well as the constellation, are shown in Table~\ref{table}.
    
\subsubsection{Downstream tasks}
The simulations are conducted on PyTorch ver. 2.0.0, CUDA ver. 11.7, and Python ver. 3.9
    \begin{itemize}
        \item \textbf{Land cover analysis on EuroSAT.~\cite{eurosat}} Land cover analysis involves the classification of the Earth's surface based on the type of vegetation, buildings, and other features in a given area. The task-specific head is a simple fully-connected network. The batch size is set to 16. We use the Adamw optimizer with a learning rate of 2e-4. 
        \item \textbf{Semantic segmentation on SegMunich.~\cite{SpectralGPT}} Semantic segmentation is to classify each pixel in an image into predefined categories. The SegMunich dataset is a collection of annotated street-level images used primarily for semantic segmentation in the context of urban environments. During the fine-tuning of the dataset, we used a batch size of 8 and set the learning rate to 1e-5. 
    \end{itemize}
\subsubsection{Baselines} To verify the effectiveness of the proposed framework, we compare it against the following algorithms:
    \begin{itemize}
       \item \textbf{Training on satellites(ToS).} This represents a traditional satellite-only federated learning paradigm where each satellite independently trains a local model using its on-board data, and task-specific head parameters are aggregated through the proposed communication strategy. It is chosen to highlight the inefficiency of on-board fine-tuning for large foundation models due to constrained computational resources. This baseline directly contrasts with our framework’s satellite-ground collaborative design, emphasizing the value of offloading heavy computations to ground servers.
        \item \textbf{Centralized training on the ground.} This baseline simulates a centralized training scheme where raw satellite data is downloaded to a ground server for model training, with no on-board processing. It is included to compare the accuracy of centralized training and federated training.
        \item \textbf{DSFL~\cite{DSFL}.} A decentralized framework where satellites collaborate via intra-orbit and inter-orbit ISLs without a central ground server.
        \item \textbf{Satellite cluster FL~\cite{10409275}.}The on-Board federated learning for satellite clusters with inter-Satellite links algorithm is a centralized framework, which uses intra-orbit links to construct a stable loop topology for low-orbit satellite constellations, and allows satellites in the same orbit to accumulate local gradients and neighbor gradients through an incremental aggregation mechanism and transmit them to the parameter server by sink nodes, and selects the appropriate sink nodes in the communication window based on predictive routing.
    \end{itemize}

    \begin{table}[]
        \centering
        \caption{Simulation parameters}
        \resizebox{1\linewidth}{!}{
        \begin{tabular}{c|c|c} \hline
             Parameter&Value&Remarks  \\ \hline
             Constellation& Walker &LEO constellation \\
             P&4&The number of orbit \\ 
             H&590 km&The height of orbits \\ 
             N&20&The number of satellites in each orbit \\
              M&80&The total number of satellites \\
             F&45 degree&Phase factor \\ 
            I&90 degree&Orbit inclination \\ 
            $T_s$&24 Sep 2020 16:00:00.00 &start time of scenario\\
            $T_e$&25 Sep 2020 16:00:00.00 &end time of scenario\\ \hline
             F&20 GHz&Transmitter frequency \\ 
             B&250 MHz&Transmitter bandwidth \\ 
             $P_t$&50 dBm&Transmitter power \\
             $G_t$&35 dBi&Transmitter gain \\ 
             $G_r$&35 dBi&Receiver gain \\ 
             $(k,\alpha)$& $(0.0751 dB/km\cdot(mm/h)^{(-\alpha)},0.083)$ & Rain attenuation coefficient \\
             \hline
        \end{tabular}
        }
        \label{table}
    \end{table}
    
\subsection{Simulation Results}
\subsubsection{Training Time} The training time of the proposed framework is compared with baselines as shown in Fig~\ref{experiment1}. It can be observed that the proposed framework reduces training time to 33\% of on-board training time. This reduction in training time can be attributed to the fact that, in the proposed framework, the majority of the computation workload is handled by the PS, which operates at a much higher speed than the computing devices of the satellites. This highlights that when the model training cost is substantial, the time required for satellites to train the entire model locally often exceeds the time it takes to transmit training data between the GSs and the satellites. This discrepancy is primarily due to the limited processing power of the satellites, which makes local training less efficient. In contrast, by offloading the heavy computation tasks to the PS and only transmitting small training information, the proposed framework reduces the overall training time, thus optimizing the overall performance of the system. 

The latency of traditional satellite federated learning is mainly dominated by computational latency, with transmission latency accounting for a very small proportion. The computing speed of satellites is much lower than that of ground servers, while the backbone network of remote sensing models requires a lot of computing power. When training on a satellite, it takes a lot of time to complete forward propagation and back propagation. The amount of updated model parameters is very small (for example, the task header parameters are only a few MB) and the transmission rate of ISLs can reach Gbps. Therefore, the impact of transmission delay on the total delay can be ignored. Therefore, the total latency in different on-board federated learning algorithms is roughly the same.

\subsubsection{Time Components}
Training time can be classified into five types. Their specific definitions, characteristics, and functions are as follows: 
\begin{itemize}
    \item On-board computation: It refers to the computing tasks executed locally on the satellite, mainly involving the computation of the embedding layer and task-specific heads.
    \item Terrestrial computing: It refers to the computing tasks executed by PS, such as the computation of the model's backbone network.
    \item Intra-Orbit transmission: Intraorbit transmission refers to data exchange among satellites within the same orbit, which functions to aggregate embedding vectors and synchronize task-specific head parameters. Within this transmission procedure, embedding vectors constitute the predominant payload, with their volume substantially exceeding that of head parameters. This discrepancy arises primarily from two key factors: first, the size of individual embedding vectors, measured in megabytes is significantly larger than that of head parameters, which are on the order of kilobytes; second, the quantity of embedding vectors is proportional to the number of data samples processed on each satellite, whereas each satellite maintains only a single task-specific head.
    \item Inter-orbit Transmission: It refers to the data transmission between satellites in different orbits, which is used for the aggregation and broadcasting of global task header parameters.
    \item Satellite-ground transmission: It refers to the data transmission between the satellite and the GS, which is used for downloading the embedding vectors and uploading the feature vectors.
\end{itemize}

\subsubsection{The Effect of Data Volume}
\begin{figure}[t]
    \centering
    \includegraphics[width=0.9\linewidth]{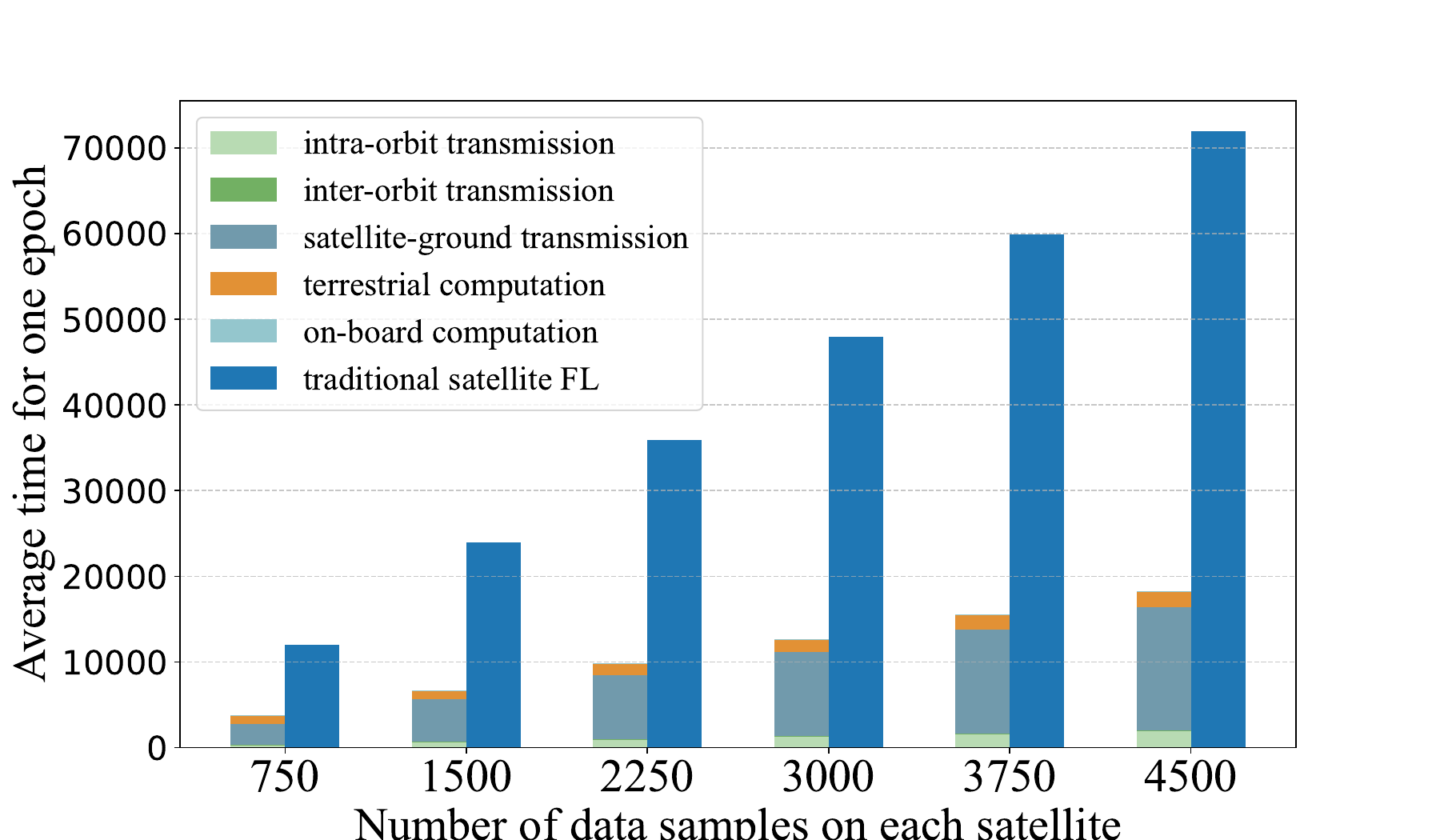}
    \caption{The effect of data volume}
    \label{dv}
\end{figure}

We analyzed the impact of the local dataset size of each satellite on time components as shown in Fig.~\ref{dv}. On-board computation mainly involves the computation of the embedding layer and task-specific heads. Its time increases linearly with the growth of sample size. This is because satellites need to perform embedding layer and loss calculations for each sample. However, the on-board computation time accounts for a small proportion due to the relatively small on-board computing tasks. The data volume also has an impact on terrestrial computing, but ground servers can alleviate the time increase caused by data volume by dynamically allocating server clusters. Intra-orbit transmission refers to data transmission between satellites within the same orbit, used for embedding vector aggregation and task header parameter synchronization. The time for embedding vector broadcasting increases linearly with the increase in sample size. This is because satellites in the same orbit need to exchange the generated embedding vectors, and even with a parallel strategy, the total transmission time still increases with the sample size. Inter-orbit transmission is less affected by the sample size, as it is used for global task head parameter aggregation. Satellite-ground transmission is used for downloading embedding vectors and uploading feature vectors. It is affected linearly by the number of samples. In summary, the time of on-board computing, intra-orbit transmission, and satellite-ground transmission increases linearly with data volume. Terrestrial computing and inter-orbit transmission are less affected by data volume. Traditional methods exhibits a longer training time and the training time in traditional methods increases linearly with data volume.

\subsubsection{The Effect of Data Rate}

The SGL data rate and ISL data rate mainly affect the total training time by influencing intra-orbit transmission, inter-orbit transmission, and satellite-ground transmission, but have no direct impact on onboard computing and terrestrial computing.

\begin{figure}[t]
    \centering
    \includegraphics[width=0.9\linewidth]{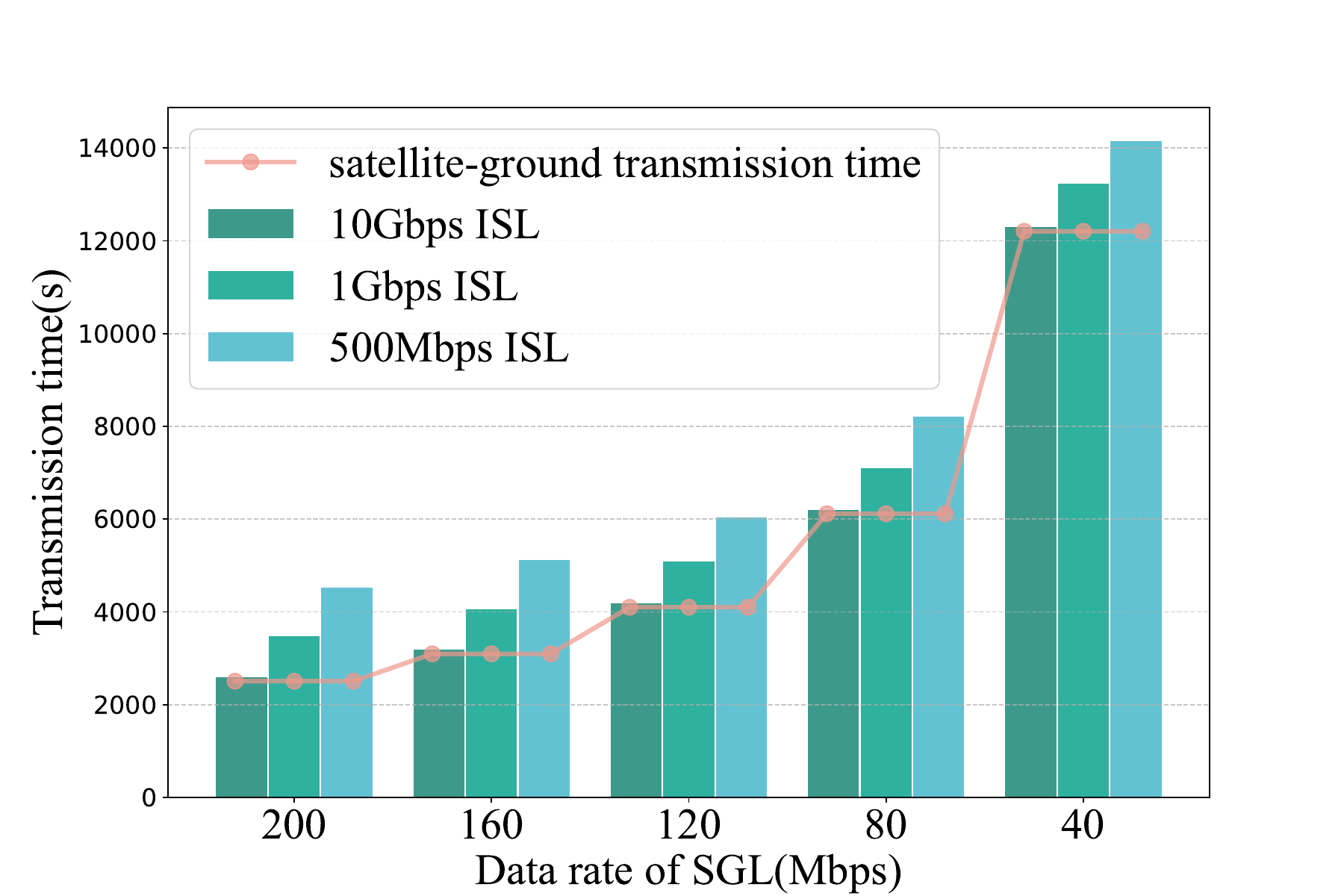}
    \caption{The effect of data rate}
    \label{dr}
\end{figure}

The SGL data rate directly determines the transmission efficiency between the satellite and the ground station, mainly affecting satellite-ground transmission, and having almost no impact on other time components. The satellite-ground transmission time taken is negatively correlated with the SGL data rate when the total data volume of embedding vectors and feature vectors is fixed. The Fig.~\ref{dr} shows that regardless of the ISL rate being 10 Gbps, 1 Gbps, or 500 Mbps, the transmission time significantly decreases as the SGL data rate increases from 40 to 200 Mbps. 

The ISL data rate influences the efficiency of data exchange between satellites, mainly affecting intra-orbit and inter-orbit transmission, and has no direct impact on satellite-to-ground transmission or computing. Intra-orbit transmission achieves embedded vector broadcasting and task header parameter aggregation through ISLs, using the Ring Allreduce parallel strategy. The time is negatively correlated with the ISL rate. Fig.~\ref{dr} shows that at the same SGL rate, the transmission time for a 10 Gbps ISL is shorter than that for a 1 Gbps and 500 Mbps ISL. Inter-orbit transmission achieves global task header parameter aggregation through inter-satellite links (ISLs), and its time is also negatively correlated with the ISL rate. When ISL data rate reach 10Gbps, satellite-ground transmission time accounts for a huge proportion of total transmission time.

\subsubsection{The Effect of Constellation Size}

\begin{figure}
    \centering
    \includegraphics[width=0.9\linewidth]{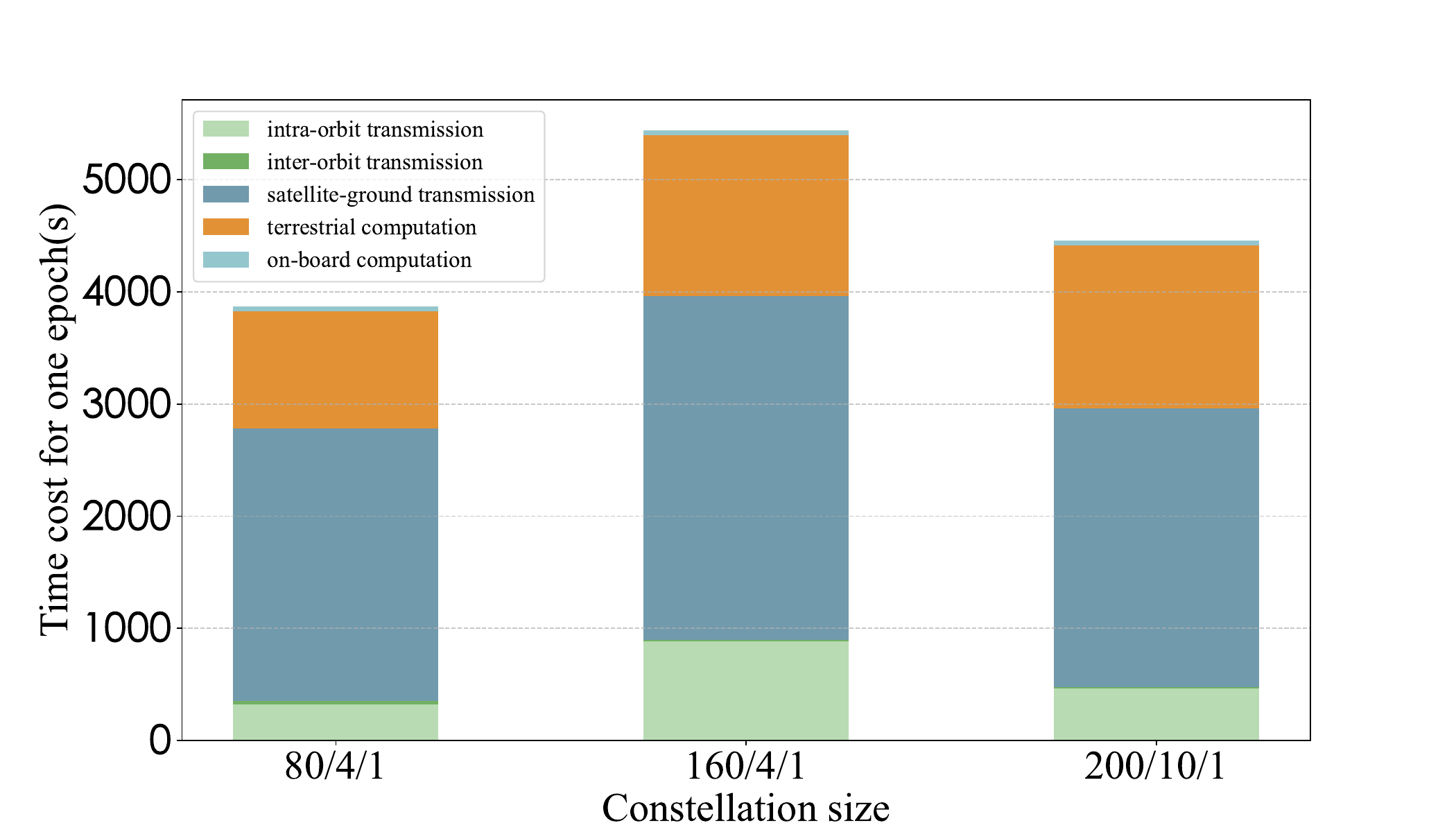}
    \caption{The effect of constellation size}
    \label{cons}
\end{figure}

When there are different number of satellites in each orbit, An increase in the number of satellites per orbit will increase the transmission time for embedding vector broadcasting, embedding vector downloading and task header parameter aggregation. Even with a parallel transmission strategy, the intra-orbit transmission time will rise due to the increased number of embedding vectors that need to be exchanged among satellites in the same orbit. In Fig.~\ref{cons}, as the number of satellites per orbit increases, the transmission time for embedding vector broadcasting shows a linear growth. An increase in the number of satellites per orbit also leads to an increase in the total data volume for satellite-ground transmission. Because more satellites lead to more connection chances, satellite-ground transmission time does not increase linearly.

An increase in the number of orbits affects the path selection of inter-orbit data transmission. When performing global task header parameter aggregation, the transmission delay is increased linearly. More orbits mean that more satellites need to communicate with the ground station, increasing the overhead of SGLs. The increase of satellite-ground transmission time is not linear because more satellites lead to more connection windows.

\subsubsection{Model Size} 
\begin{figure}
    \centering
    \includegraphics[width=0.9\linewidth]{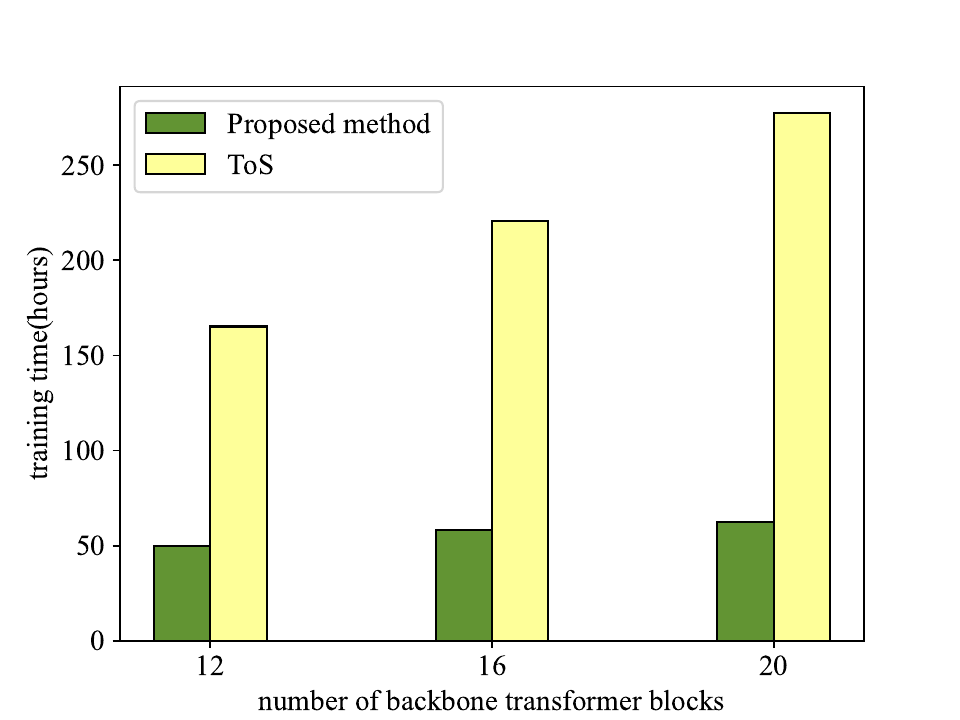}
    \caption{Total training time versus model size.}
    \label{model_size}
    \vspace{-0.5cm}
\end{figure}

We also analyze the impact of different model sizes on the training efficiency of the proposed framework. The backbone network of our model is composed of several transformer blocks, and we conduct simulations by varying the number of these blocks. Specifically, these simulations are performed on the EuroSAT dataset, and the results are depicted in Fig.~\ref{model_size}. 
The simulation results indicate that as the model size increases, the training time gap between our proposed framework and the traditional on-board training approach becomes progressively larger. 
The underlying reason is the increased computation demand in the backbone network as more transformer blocks are added. 
As the model size increases, a larger portion of the overall training process is accelerated on the PS, while the volume of data transmitted between the GSs and the satellites remains constant. This imbalance between local computation capacity and the model's demands amplifies the advantage of offloading the computation to the GS, leading to a larger reduction in training time. This effect becomes even more pronounced as the model size increases, highlighting the scalability of the proposed framework.

\subsubsection{The Overhead of Satellite-Ground Transmission}
Intermittent SGLs can prevent data from being downloaded to the ground server in central training. To assess whether the proposed framework can effectively mitigate these issues, we conduct an analysis of the communication overhead. Specifically, we compare the communication overhead in our framework with the traditional approach of downloading the entire original dataset. For this comparison, we use several commonly used datasets in RS applications that cover a variety of domains, such as environmental monitoring and geospatial analysis. We simulate the data transmission requirements for both the traditional method of downloading the full dataset and our proposed framework that focuses on transmitting only model updates rather than raw data.
\begin{enumerate}
    \item Bridge dataset~\cite{bridge}. 
    This dataset is for object detection and consists of 500 images, each containing at least one bridge from different regions worldwide. All images are $4,800 \times 2,843$ pixels.
    \item Aerial image segmentation dataset~\cite{Aerial}. 
    This is used in image segmentation, where a semantic class label is assigned to each pixel as a basis for automatic map generation. Each image is approximately $3000 \times 3000$ pixels.
    \item OSCD~\cite{OSCD}. It contains 24 pairs of multispectral images taken between 2015 and 2018 and provides pairs of satellite images in 13 bands that can be used for change detection. The spatial resolution of the image varies between 10m, 20m, and 60m. The image size is $600 \times 600$ pixels.
    \item SegMunich dataset~\cite{SpectralGPT}. It is used in previous simulations.
\end{enumerate}
 
\begin{figure}
    \centering
    \includegraphics[width=0.9\linewidth]{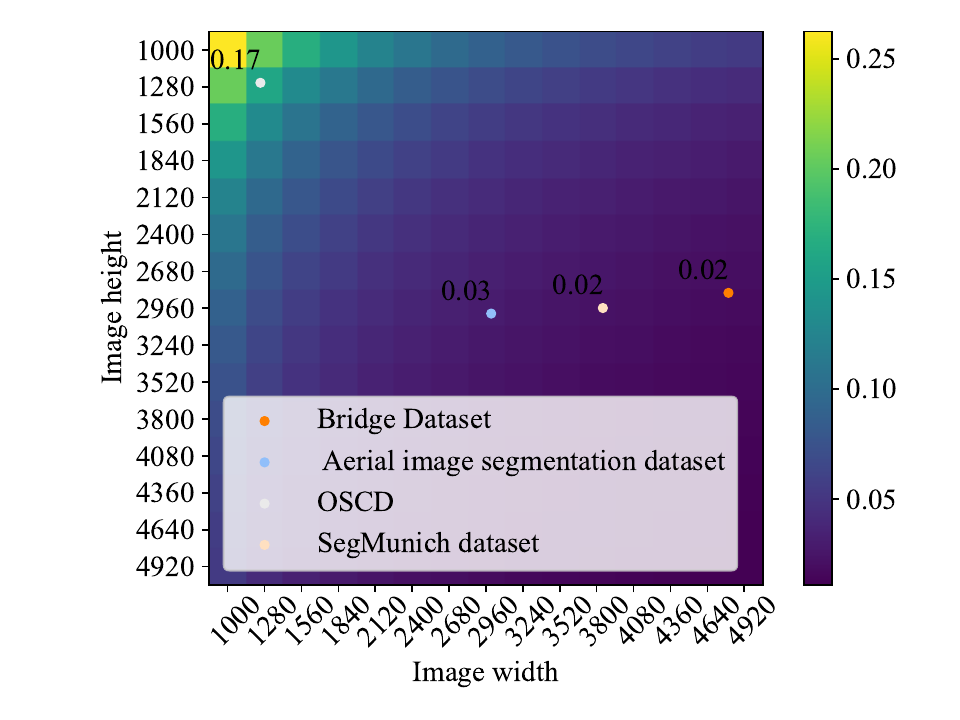}
    \caption{Communication overhead of SGLs. We show the ratio of the transmitted data size to the original image size (width $\times$ height) in this figure. }
    \label{data_size}
    \vspace{-0.3cm}
\end{figure}

 We present the results in Fig.~\ref{data_size}. For datasets with 13 bands, we convert them into images with 3 bands of the same size to display them in the figure. The figure displays the ratio of the transmitted data size to the original data size when the original images vary in size. This ratio provides a clear indication of how much data is being transmitted between the satellite and the GS relative to the size of the original dataset. As the size of the original images increases, this ratio decreases, which can be attributed to the information extraction capability of the FMs. 
 Several points in the figure are marked, each representing a specific scenario where the data transmission ratio is calculated for a given dataset. For example, in the simulations conducted on the SegMunich dataset, the transmitted data size is approximately $2\%$ of the size of the original dataset. 
 
 Given device parameters set as in Table~\ref{table}, the estimated data rate for downlinks is approximately $200\text{Mbps}$, according to formulas in Section~\ref{communication network}. This data rate is insufficient for transmitting large volumes of raw data in a constellation. 
 Consequently, directly transmitting raw data from the satellite to the ground server would result in significant delays, increased latency, and system performance bottlenecks. The proposed framework reduces the amount of data transmitted to the ground server, which alleviates the pressure on intermittent SGLs. Therefore, in certain scenarios where satellites continuously collect high-precision images daily and the data generation speed exceeds the transmission speed, the proposed framework proves more effective than traditional methods of downloading raw data. 

\subsubsection{Communication Strategy} To verify the effective co-design of computing and communication in the proposed framework, we apply various inter-satellite and satellite-ground communication strategies.

\begin{enumerate}
    \item \textbf{Strategy 1:} Sequential intra-orbit transmission. In this strategy, intra-orbit data are first collected and aggregated by a designated satellite. Then, the satellite sends the collected data to other satellites in the same orbit via ISLs. Namely, data are transmitted sequentially in a specified direction until all satellites obtain $\boldsymbol{E}_p$. The total time required for this process is given by:
    \begin{align}
    T = 2 \cdot {N} \cdot c,
    \end{align}
    where $c$ represents the time required to transmit data between two adjacent satellites. It can be observed that the total time increases linearly with the number of satellites.
    \item  \textbf{Strategy 2:} Satellite-ground transmission without inter-satellite cooperation(Satellite-ground transmission without ISC)~\cite{9674028}. Each satellite only transmits local data until the PS receives all data.
    \item  \textbf{Strategy 3:} Inter-orbit gossip transmission~\cite{DFedSat}. Each orbit only receives model parameters from two adjacent orbits and repeats the process for several rounds.
\end{enumerate}

We can find that sequential intra-orbit transmission significantly reduces convergence speed compared to parallel transmission.  In sequential transmission, after the computation is performed on each satellite, the data is transmitted sequentially. At any given moment, only one satellite is actively transmitting its data to the other satellite, while all other satellites remain idle. As a result, the overall system throughput is reduced, and the communication link is underutilized. Strategy 2 prolongs training time due to stragglers. In this strategy, the PS is required to wait for the slowest satellites to finish transmitting their model updates before starting the next training epoch. This creates a significant delay. For strategy 3, inter-orbit gossip transmission has a similar training time in each training round but needs more epochs to converge because all satellites do not reach a consensus on the global model in each epoch. These simulations demonstrate that the co-design of computing and communication plays a pivotal role in reducing latency, minimizing communication overhead, and accelerating the training process. 

\begin{figure}[t]
    \centering
    \subfloat[Test accuracy]{
        \includegraphics[width=0.8\linewidth]{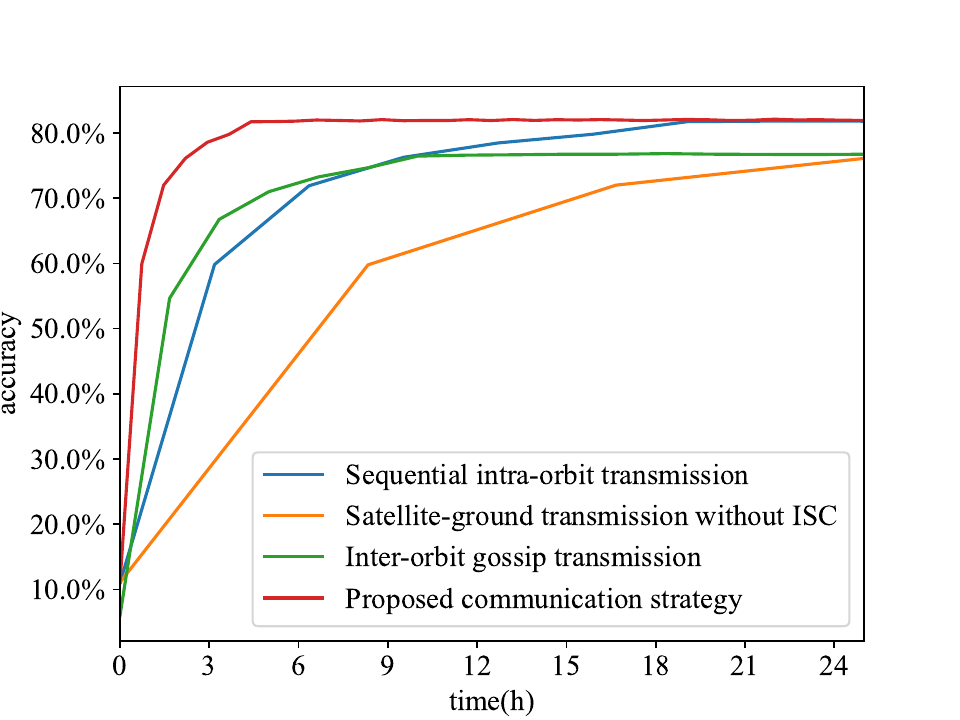}
        }
        \\
    \subfloat[Training loss]{
        \includegraphics[width=0.8\linewidth]{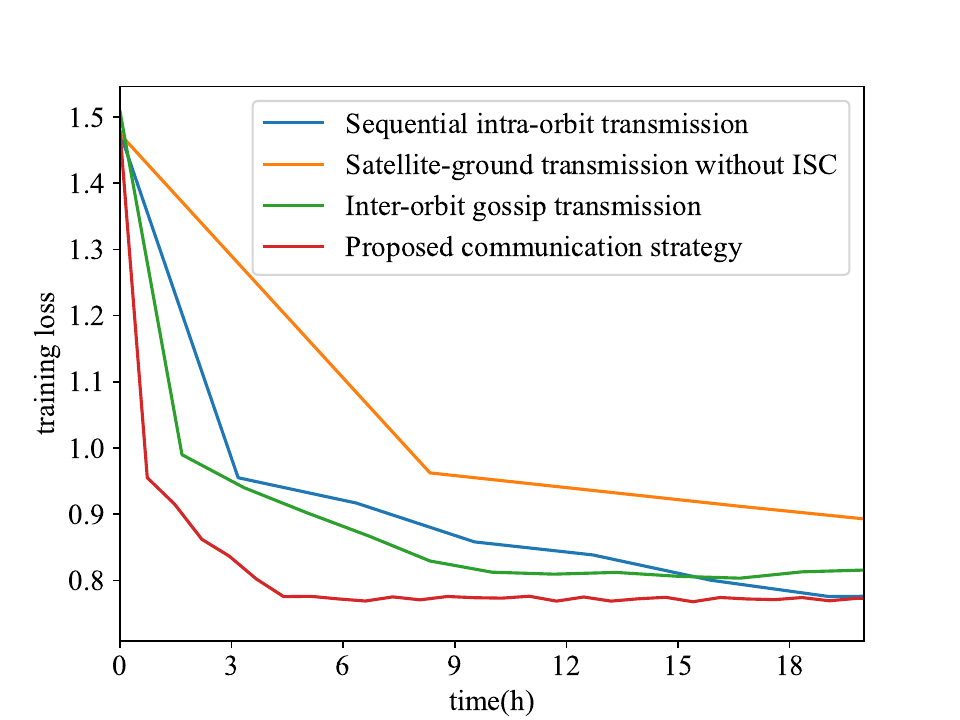}
        }
    \caption{Learning performance versus training time in different strategies.}
    \label{ex4}
    \vspace{-0.6cm}
\end{figure}
\subsubsection{Link Utilization}

The satellite-ground communication network is modeled as a directed acyclic graph with capacity constraints, and the maximum flow algorithm is used to allocate transmission tasks. In each time slice, all available SGLs are used in parallel and fully fill the communication window. And the embedding vectors aggregated in the orbit are transmitted simultaneously for satellites. For example, after satellites in a certain orbit aggregate data through a ring broadcast, it can be transmitted by any satellite in the communication window, without the need for other satellites to wait, thus reducing the link idle time and improve the link utilization.

During intra-orbit transmission, the data is divided into N blocks. In the scatter-reduce phase, each satellite sends data to its neighbors and receives  data from its neighbors, completing local aggregation synchronously. In the collection phase, the aggregated data blocks are transmitted in reverse. During the entire process, all links are occupied at the same time, and there are no idle links, and the link utilization rate is close to 100\%.

\section{Conclusion}
\label{conclusion}
This paper proposed a satellite federated fine-tuning framework for RS FMs to addresses the training latency and computation resource constraints stemming from the limited computational capabilities of satellites, intermittent SGLs, and dynamic ISLs in Space-CPN. To tackle these issues, we designed a model partitioning and training scheme. We partition models into the embedding layer, the backbone network and the output layer and deploy the backbone networks and other components on the PS and satellites respectively to reduce on-board computation load and transmission burden. Furthermore, communication strategies for both satellite-ground and inter-satellite transmissions were co-designed with the training process to minimize latency. We put forward a topology-aware communication scheme, which assigns transmission tasks according to the real-time topological structure and the status of links. Regarding intra-orbit communication, considering the ring topology of each orbit, we present a parallel communication method founded on Ring Allreduce. As for inter-orbit communication, we design a communication algorithm that aims to reduce latency to the minimum by taking link capacity into account. These strategies blend communication and computational procedures, factoring in aspects such as restricted bandwidth, sparse connectivity, and dynamic topologies. As a result, these strategies enhance the efficiency of data transmission and expedite the convergence of models. Simulation results validate the effectiveness of our proposed framework and communication strategies.

\bibliographystyle{IEEEtran}
\bibliography{IEEEabrv}

\end{document}